\documentclass{article} 
\usepackage{iclr2022_conference,times}

\usepackage{amsmath,amsfonts,bm}

\def\eqref#1{equation~\ref{#1}}

\def\1{\bm{1}}

\DeclareMathAlphabet{\mathsfit}{\encodingdefault}{\sfdefault}{m}{sl}
\SetMathAlphabet{\mathsfit}{bold}{\encodingdefault}{\sfdefault}{bx}{n}

\usepackage[colorlinks=true]{hyperref}
\usepackage{url}
\usepackage{amsmath}
\usepackage{multirow}
\usepackage{booktabs}
\usepackage{graphicx}
\usepackage{subcaption}
\usepackage{wrapfig}
\iclrfinalcopy 
\hypersetup{
  colorlinks,
  citecolor=black,}

\definecolor{myRed}{rgb}{0.8, .2, .2}

\title{Quadtree Attention for Vision Transformers}

\author{Shitao Tang$^{1\ast}$, Jiahui Zhang$^{2}$\thanks{Equal contribution}, Siyu Zhu$^{2}$, Ping Tan$^{12}$ \\
$^{1}$Simon Fraser University, $^{2}$Alibaba A.I. Lab\\
\texttt{shitaot@sfu.ca}, \texttt{zjhthu@gmail.com},\\
\texttt{siting.zsy@alibaba-inc.com}, \texttt{pingtan@sfu.ca} \\
}

\iclrfinalcopy 
\begin{document}

\maketitle

\begin{abstract}
Transformers have been successful in many vision tasks, thanks to their capability of capturing long-range dependency. However, their quadratic computational complexity poses a major obstacle for applying them to vision tasks requiring dense predictions, such as object detection, feature matching, stereo, etc. We introduce QuadTree Attention, which reduces the computational complexity from quadratic to linear. Our quadtree transformer builds token pyramids and computes attention in a coarse-to-fine manner. At each level, the top $K$ patches with the highest attention scores are selected, such that at the next level, attention is only evaluated within the relevant regions corresponding to these top $K$ patches. We demonstrate that quadtree attention achieves state-of-the-art performance in various vision tasks, e.g. with 4.0\% improvement in feature matching on ScanNet, about 50\% flops reduction in stereo matching, 0.4-1.5\% improvement in top-1 accuracy on ImageNet classification, 1.2-1.8\% improvement on COCO object detection, and 0.7-2.4\% improvement on semantic segmentation over previous state-of-the-art transformers. The codes are available at \href{https://github.com/Tangshitao/QuadtreeAttention}{https://github.com/Tangshitao/QuadtreeAttention}.
\end{abstract}

\section{Introcution}

Transformers can capture long-range dependencies by the attention module and have demonstrated tremendous success in natural language processing tasks. In recent years, transformers have also been adapted to computer vision tasks for  image classification ~\citep{dosovitskiy2020image}, object detection~\citep{wang2021pyramid}, semantic segmentation~\citep{liu2021swin}, feature matching~\citep{sarlin2020superglue}, and stereo~\citep{li2020revisiting}, etc. 
Typically, images are divided into patches and these patches are flattened and fed to a transformer as word tokens to evaluate attention scores. 
However, transformers have quadratic computational complexity in terms of the number of tokens, i.e. number of image patches. Thus, applying transformers to computer vision applications requires careful simplification of the involved computation.

To utilize the standard transformer in vision tasks, many works opt to apply it on low resolution or sparse tokens. ViT~\citep{dosovitskiy2020image} uses coarse image patches of $16\times 16$ pixels to limit the number of tokens. DPT~\citep{ranftl2021vision} up-samples low-resolution results from ViT to high resolution maps to achieve dense predictions. 
SuperGlue~\citep{sarlin2020superglue} applies transformer on sparse image keypoints. 
Focusing on correspondence and stereo matching applications, ~\cite{germain2021visual} and~\cite{li2020revisiting} also apply transformers at a low resolution feature map. 



%



However, as demonstrated in several works~\citep{wang2021pyramid,liu2021swin, sun2021loftr, li2020revisiting, shao2020channel}, applying transformers on high resolution is beneficial for a variety of tasks. Thus, many efforts have been made to design efficient transformers to reduce computational complexity. 
Linear approximate transformers~\citep{katharopoulos2020transformers, wang2020linformer} approximate standard attention computation with linear methods.
However, empirical studies~\citep{germain2021visual, chen2021learning} show those linear transformers are inferior in vision tasks. To reduce the computational cost, the PVT~\citep{wang2021pyramid} uses downsampled keys and values, which is harmful to capture pixel-level details. In comparison, the Swin Transformer~\citep{liu2021swin} restricts the attention in local windows in a single attention block, which might hurt long-range dependencies, the most important merit of transformers. 



\begin{figure}
    \centering
    \includegraphics[width=0.8\textwidth]{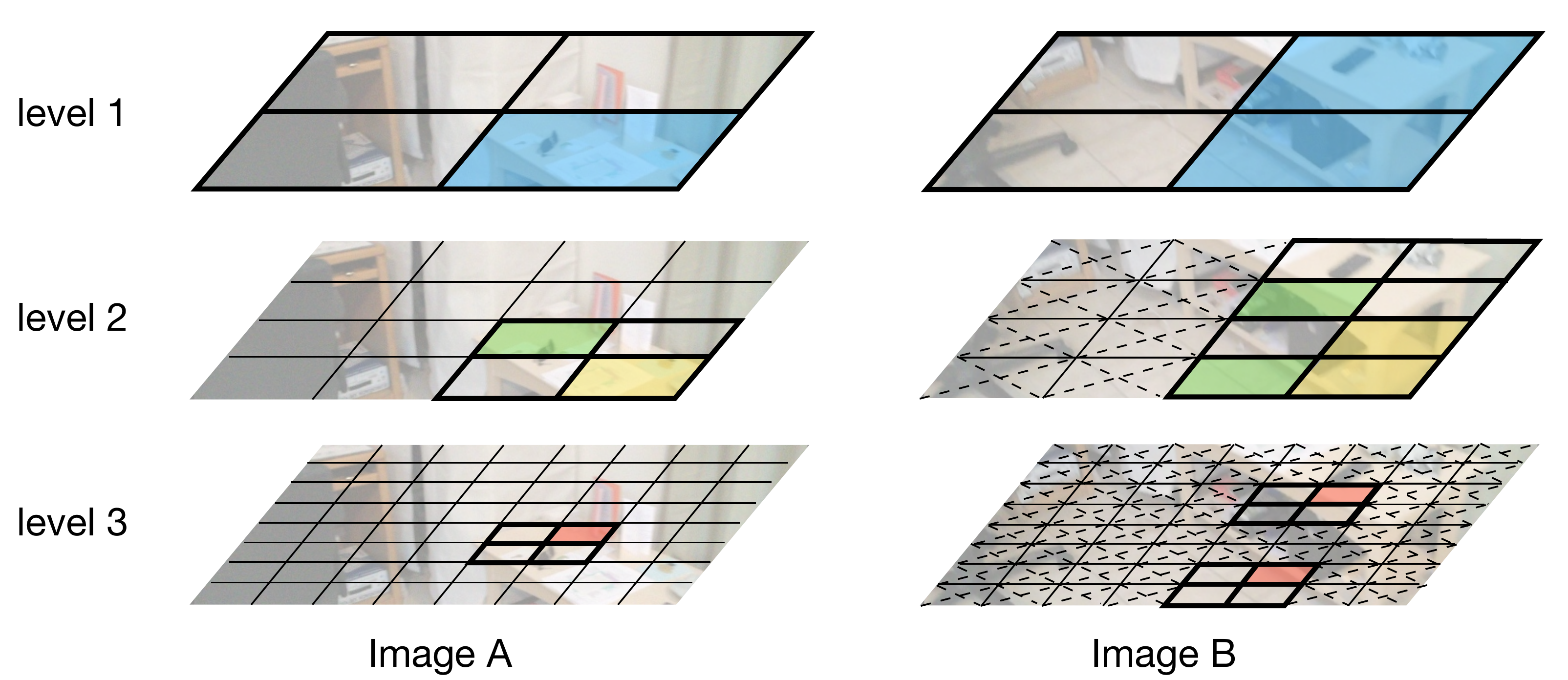}
    \caption{Illustration of QuadTree Attention. Quadtree attention first builds token pyramids by down-sampling the query, key and value. From coarse to fine, quadtree attention selects top $K$ (here, $K=2$) results with the highest attention scores at the coarse level. At the fine level, attention is only evaluated at regions corresponding to the top $K$ patches at the previous level. The query 
   sub-patches in fine levels share the same top $K$ key tokens and coarse level messages, e.g., green and yellow sub-patches at level 2 share the same messages from level 1. We only show one patch in level 3 for simplicity.}
    \label{fig:quadtree}
    \vspace{-6mm}
\end{figure}

Unlike all these previous works, we design an efficient vision transformer that captures both fine image details and long-range dependencies. 
Inspired by the observation that most image regions are irrelevant, we build token pyramids and compute attention in a coarse to fine manner. In this way, we can quickly skip irrelevant regions in the fine level if their corresponding coarse level regions are not promising.
For example, as in Figure~\ref{fig:quadtree}, at the 1st level, we compute the attention of the blue image patch in image A with all the patches in image B and choose the top $K$ (here, $K=2$) patches which are also highlighted in blue. In the 2nd level, for the four framed sub-patches in image A (which are children patches of the blue patch at the 1st level), we only compute their attentions with the sub-patches corresponding to the top $K$ patches in image B at the 1st level. All the other shaded sub-patches are skipped to reduce computation. We highlight two sub-patches in image A in yellow and green. Their corresponding top $K$ patches in image B are also highlighted in the same color. This process is iterated in the 3rd level, where we only show the sub-sub-patches corresponding to the green sub-patch at the 2nd level. 
In this manner, our method can both obtain fine scale attention and retain long-range connections. Most importantly, only sparse attention is evaluated in the whole process. Thus, our method has low memory and computational costs. Since a quadtree structure is formed in this process, we refer to our method as QuadTree Attention, or QuadTree Transformer.

In experiments, we demonstrate the effectiveness of our quadtree transformer in both tasks requiring cross attention, e.g. feature matching and stereo, and tasks only utilizing self-attention, e.g. image classification and object detection. Our method achieves state-of-the-art performance with significantly reduced computation, comparing to relevant efficient transformers (\cite{katharopoulos2020transformers, wang2021pyramid, liu2021swin}).
In feature matching, we achieve 61.6 AUC@20$^{\circ}$ in ScanNet~\citep{dai2017scannet}, 4.0 higher than the linear transformer~\citep{katharopoulos2020transformers} but with similar flops. 
In stereo matching, we achieve a similar end-point-error as standard transformer, \citep{li2020revisiting} but with about 50\% flops reduction and 40\% memory reduction.
In image classification, we achieve 84.0\% top-1 accuracy in ImageNet~\citep{deng2009imagenet}, 5.7\% higher than ResNet152~\citep{he2016deep} and 1.0\% higher than the Swin Transformer-S~\citep{liu2021swin}. In object detection, our QuadTree Attention + RetinaNet achieves 47.9 AP in COCO~\citep{lin2014microsoft}, 1.8 higher than the backbone PVTv2~\citep{wang2021pvtv2} with fewer flops. In semantic segementation, QuadTree Attention improves the performance by 0.7-2.4\%.


\section{Related work}

\textbf{Efficient Transformers.} Transformers have shown great success in both natural language processing and computer vision. Due to the quadratic computational complexity, the computation of full attention is unaffordable when dealing with long sequence tokens. Therefore, many works design efficient transformers, aiming to reduce computational complexity~\citep{katharopoulos2020transformers, choromanski2020rethinking,shao2021dynamic, wang2020linformer, lee2019set, ying2018hierarchical}. Current efficient transformers can be categorized into three classes. 1) Linear approximate attention~\citep{katharopoulos2020transformers, choromanski2020rethinking, wang2020linformer, beltagy2020longformer, zaheer2020big} approximates the full attention matrix by linearizing the softmax attention and thus can accelerate the computation by first computing the product of keys and values.
2) Inducing point-based linear transformers ~\citep{lee2019set, ying2018hierarchical} use learned inducing points with fixed size to compute attention with input tokens, thus can reduce the computation to linear complexity.
However, these linear transformers are shown to have inferior results than standard transformers in different works ~\citep{germain2021visual,chen2021learning}.
3) Sparse attention, including Longformer~\citep{beltagy2020longformer}, Big Bird~\citep{zaheer2020big}, etc, attends each query token to part of key and value tokens instead of the entire sequence. Unlike these works, our quadtree attention can quickly skip the irrelevant tokens according to the attention scores at coarse levels. Thus, it achieves less information loss while keeps high efficiency.

\textbf{Vision Transformers.} Transformers have shown extraordinary performance in many vision tasks. ViT~\citep{dosovitskiy2020image} applies transformers to image recognition, demonstrating the superiority of transformers for image classification at a large scale. However, due to the computational complexity of full attention, it is hard to apply transformers in dense prediction tasks, e.g. object detection, semantic segmentation, etc. To address this problem, Swin Transformer~\citep{liu2021swin} restricts attention computation in a local window. Focal transformer~\citep{yang2021focal} uses two-level windows to increase the ability to capture long-range connection for local attention methods.
Pyramid vision transformer (PVT)~\citep{wang2021pyramid} {reduce the computation of global attention methods by downsampling key and value tokens.}
Although these methods have shown improvements in various tasks, they have drawbacks either in capturing long-range dependencies ~\citep{liu2021swin} or fine level attention ~\citep{wang2021pyramid}.
Different from these methods, our method simultaneously capture both local and global attention by computing attention from full image levels to the finest token levels with token pyramids in one single block.
Besides, the K-NN transformers~\citep{wang2021kvt, zhao2019explicit} aggregate messages from top $K$ most similar tokens as ours, but they compute the attention scores among all pairs of query and key tokens, and thus still has quadratic complexity.

Beyond self-attention, many tasks can largely benefit from cross attention. Superglue~\citep{sarlin2020superglue} processes detected local descriptors with self- and cross attention and shows significant improvement in feature matching. Standard transformers can be applied in SuperGlue because only sparse keypoints are considered. SGMNet ~\citep{chen2021learning} further reduces the computation by attending to seeded matches. LoFTR~\citep{sun2021loftr} utilizes linear transformer~\citep{katharopoulos2020transformers} on low-resolution feature maps to generate dense matches.
For stereo matching, STTR ~\citep{li2020revisiting} applies self- and cross attention along epipolar lines and reduces the memory by gradient checkpointing engineering techniques.
However, due to the requirement of processing a large number of points, these works either use linear transformers, which compromise performance, or a standard transformer, which compromises efficiency. In contrast, our transformer with quadtree attention achieves a significant performance boost compared with linear transformer or efficiency improvement compared with standard transformer. Besides, it can be applied to both self-attention and cross attention.

\section{Method}
We first briefly review the attention mechanism in transformers in Section~\ref{sec:attn} and then  formulate our quadtree attention in Section~\ref{sec:quadtree_att}.

\subsection{Attention in Transformer}\label{sec:attn}
Vision transformers have shown great success in many tasks.
At the heart of a transformer is the attention module, which can capture long-range information between feature embeddings. 
Given two image embeddings $\mathbf{X}_1$ and $\mathbf{X}_2$, the attention module passes information between them. Self-attention is the case when $\mathbf{X}_1$ and $\mathbf{X}_2$ are the same, while cross attention covers a more general situation when $\mathbf{X}_1$ and $\mathbf{X}_2$ are different.
It first generates the query $\mathbf{Q}$, key $\mathbf{K}$, and value $\mathbf{V}$ by the following equation,
\begin{align*} 
\mathbf{Q}&=\mathbf{W}_q\mathbf{X}_1, \\ 
\mathbf{K}&=\mathbf{W}_k\mathbf{X}_2,\\
\mathbf{V}&=\mathbf{W}_v\mathbf{X}_2,
\end{align*}
where $\mathbf{W}_q$, $\mathbf{W}_k$ and $\mathbf{W}_v$ are learnable parameters. Then, it performs message aggregation 
by computing the attention scores between query and key as following,
\begin{equation}
    \mathbf{Y}=\text{softmax}(\frac{\mathbf{Q}\mathbf{K}^T}{\sqrt{C}}) \mathbf{V},
    \label{equ:standard_transformer}
\end{equation}
where $C$ is the embedding channel dimension.
The above process has $O(N^2)$ computational complexity, where $N$ is the number of image patches in a vision transformer. This quadratic complexity hinders transformers from being applied to tasks  requiring high resolution output. To address this problem, PVT (\cite{wang2021pyramid}) downsamples $\mathbf{K}$ and $\mathbf{V}$, while Swin Transformer (\cite{liu2021swin}) limits the attention computation within local windows.

\begin{figure}
\centering
    \includegraphics[width=0.95\textwidth]{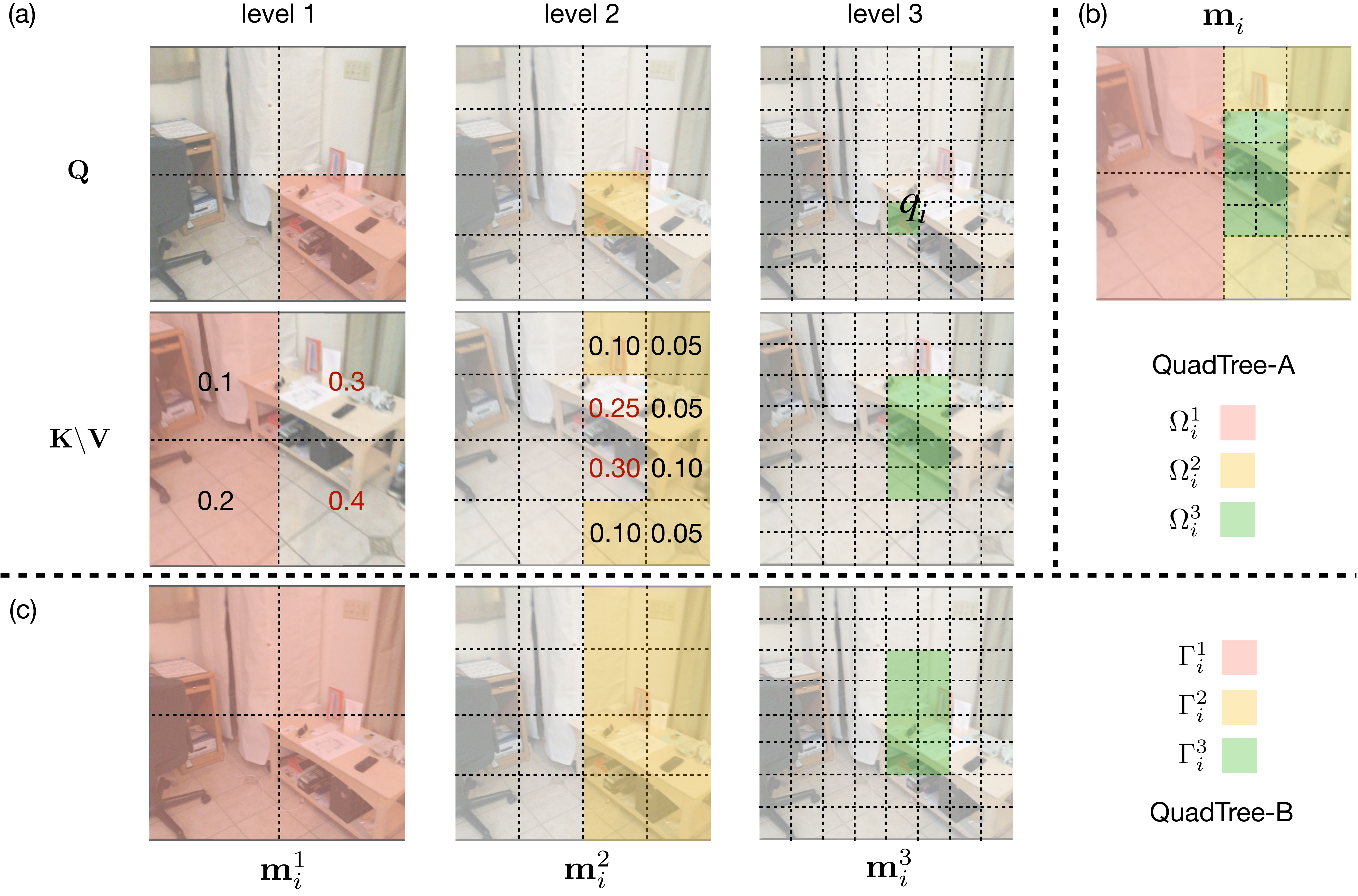}
  \caption{Illustration of quadtree message aggregation for a query token $q_i$. (a) shows the token pyramids and involved key/value tokens in each level. Attention scores are marked in the first two levels for clarification, and the top $K$ scores are highlighted in red. (b) shows message aggregation for QuadTree-A architecture. The message is assembled from different levels along a quadtree. (c) shows message aggregation for QuadTree-B architecture. The message is collected from overlapping regions from different levels.}
    \label{fig:quadtree_message}
\end{figure}

\subsection{QuadTree Attention}\label{sec:quadtree_att}
In order to reduce the computational cost of vision transformers, we present QuadTree Attention. As the name implies, we borrow the idea from quadtrees, which are often used to partition a two-dimensional space by recursively subdividing it into four quadrants or regions. Quadtree attention computes attention in a coarse to fine manner. According to the results at the coarse level, irrelevant image regions are skipped quickly at the fine level. This design achieves less information loss while keeping high efficiency.

The same as the regular transformers, we first linearly project $\mathbf{X}_1$ and $\mathbf{X}_2$ to the query, key, and value tokens. To facilitate fast attention computation, we construct $L$-level pyramids for query $\mathbf{Q}$, key $\mathbf{K}$, and value $\mathbf{V}$ tokens by downsampling feature maps. For query and key tokens, we use average pooling layers. For value tokens, average pooling is used for cross attention tasks and convolutional-normalization-activation layers with stride 2 are used for self attention tasks if no special statement.
As shown in Figure~\ref{fig:quadtree}, after computing attention scores in the coarse level, for each query token, we select the top $K$ key tokens with the highest attention scores. At the fine level, query sub-tokens only need to be evaluated with those key sub-tokens that correspond to one of the selected $K$ key tokens at the coarse level. 
This process is repeated until the finest level. After computing the attention scores, we aggregate messages at all levels, where we design two architectures named as \textbf{QuadTree-A} and \textbf{QuadTree-B}.






\textbf{QuadTree-A}. 
Considering the $i$-th query token $\mathbf{q}_i$ at the finest level, we need to compute its received message $\mathbf{m}_i$ from all key tokens. This design assembles the full message by collecting partial messages from different pyramid levels. Specifically, 
\begin{equation}
    \mathbf{m}_i = \sum_{1\leq l \leq L} \mathbf{m}_i^l,
\end{equation}
where $\mathbf{m}_i^l$ indicates the partial message evaluated at level $l$. This partial message $\mathbf{m}_i^l$ assemble messages
at the $l$-th level from tokens within the region $\Omega_i^l$, which will be defined later. In this way, messages from less related regions are computed from coarse levels, while messages from highly related regions are computed in fine levels. This scheme is illustrated in Figure~\ref{fig:quadtree_message} (b),
message $\mathbf{m}_i$ is generated by assembling three partial messages that are computed from different image regions with different colors, which collectively cover the entire image space. The green region indicates the most relevant region and is evaluated at the finest level, while the red region is the most irrelevant region and is evaluated at the coarsest level. The region $\Omega_i^l$ can be defined as $\Gamma_i^{l} - \Gamma_i^{l+1}$, where the image region $\Gamma_i^l$ corresponds to the top $K$ tokens at the level $l-1$.  The regions $\Gamma_i^l$ are illustrated in  Figure~\ref{fig:quadtree_message} (c). The region $\Gamma_i^1$ covers the entire image.

The partial messages are computed as,
\begin{equation}
    \mathbf{m}_i^l = \sum _{j \in \Omega_i^l} s_{ij}^l \mathbf{v}_j^{l},
    \label{equ:att_score}
\end{equation}
where $s_{ij}^l$ is the attention score between the query and key tokens at level $l$. {Figure~\ref{fig:quadtree_message} (a) highlights query and key tokens involved in computing $\mathbf{m}_i^l$ with the same color as $\Omega_i^l$.}
Attention scores are computed recursively,
\begin{equation}
    s_{ij}^l = s_{ij}^{l-1} t_{ij}^l.
\end{equation}
Here, $s_{ij}^{l-1}$ is the score of corresponding parent query and key tokens and $s_{ij}^1 = 1$. The tentative attention score $t_{ij}^l$ is evaluated according to  Equation~\ref{equ:standard_transformer} among the $2\times 2$ tokens of the same parent query token. 
For QuadTree-A, we use average pooling layers to downsample all query, key and value tokens.


\textbf{QuadTree-B}. 
The attention scores $s_{ij}^l$ in \textbf{QuadTree-A} are recursively computed from all levels, which makes scores smaller at finer levels and reduces the contributions of fine image features. Besides,  fine level scores are also largely affected by the inaccuracy at coarse levels.
So we design a different scheme, referred as \textbf{QuadTree-B} in this paper, to address this problem. Specifically, we compute $\mathbf{m}_i$ as a weighted average of the partial messages from different levels, 
\begin{equation}
    \mathbf{m}_i = \sum_{1 \leq l \leq L} w_i^l \mathbf{m}_i^l,
\end{equation}
where $w_i^l$ is a learned weight. As shown in Figure~\ref{fig:quadtree_message} (c), the partial messages here overlap with each other, which are computed as,
\begin{equation}
    \mathbf{m}_i^l=\text{Attention}(\mathbf{q}_i^l, \mathbf{K}^l_{\Gamma _i^{l}}, \mathbf{V}^l_{\Gamma_i^{l}}),
    \label{equ:qudatreeB}
\end{equation}
where $\text{Attention}$ is the attention message computation as Equation~\ref{equ:standard_transformer}. Here, $\mathbf{K}^l_{\Gamma _i^{l}}$ and $\mathbf{V}^l_{\Gamma _i^{l}}$ are matrices formed by stacking all keys and values within the region $\Gamma_i^{l}$. 




Both \textbf{QuadTree-A} and \textbf{QuadTree-B} involve only sparse attention evaluation. Thus, our method largely reduces computational complexity. As analyzed in Appendix~\ref{apx:complexity}, the computational complexity of our quadtree attention is linear to the number of tokens.

   
\textbf{Multiscale position encoding}. The computation of attention is permutation invariant to tokens, and thus positional information is missed. 
To address this problem, we adopt the locally-enhanced positional encoding (LePE)~\citep{dong2021cswin} at each level to 
design a multiscale position encoding.
Specifically, for level $l$, we apply unshared depth-wise convolution layers to value tokens $\mathbf{V}^l$ to encode the positional information. \vspace{-6mm}


\section{Experiment}
We experiment our quadtree transformer with four representative tasks, including feature matching, stereo, image classification, and object detection. The first two tasks require cross attention to fuse information across different images, while the latter two involve only self-attention. We implement our quadtree transformer using PyTorch and CUDA kernels. More implementation details are provided in Appendix~\ref{sec:task_intro}.

\subsection{Cross attention tasks}\label{sec:3dvision}


\subsubsection{Feature matching}
Finding feature correspondence \citep{luo2019contextdesc,detone2018superpoint} across different images is a precedent problem for many 3D computer vision tasks. It is typically evaluated by the accuracy of the camera pose estimated from the corresponding points. We follow the framework proposed in a recent state-of-the-art work LoFTR~\citep{sun2021loftr}, which consists of a CNN-based feature extractor and a transformer-based matcher. We replace the linear transformer~\citep{katharopoulos2020transformers} in LoFTR with our quadtree transformer. Besides, we also implement a new version of LoFTR with the spatial reduction (SR) attention~\citep{wang2021pyramid} for additional comparison.

\textbf{Setting.} 
We experiment on ScanNet~\citep{dai2017scannet} with 1,513 scans. In order to accelerate training, we design the LoFTR-lite setting, which uses half of the feature channels of LoFTR and 453 training scans.
Ablation studies in section \ref{sec:analysis} are conducted in this setting.
We train both LoFTR-lite and LoFTR for 30 epochs with batch size 8. 
For quadtree transformer, we build pyramids of three levels with the coarsest resolution at $15\times 20$ pixels.
We set the parameter $K$ to 8 at the finest level, and double it at coarser levels. For the SR attention, we average pool the value and key tokens to the size $8\times 8$ to keep similar memory usage and flops as our quadtree attention. More details are included in Appendix~\ref{appendix:image match}. 


\begin{table}[]
\small
\centering
\begin{tabular}{l|l|lll}
                                           &                   & AUC@5$^{\circ}$                              & AUC@10$^{\circ}$                             & AUC@20$^{\circ}$                             \\ \hline
\multirow{4}{*}{Others}                    
                                           & ContextDesc + SGMNet(\cite{chen2021learning})           & 15.4      &32.3                                                & 48.8                        \\
                                           & SuperPoint + OANet (\cite{zhang2019learning})         & 11.8                           & 26.9                           & 43.9                           \\
                                           & SuperPoint + SuperGlue (\cite{sarlin2020superglue})         & \textbf{16.2}                          & \textbf{33.8}                           & \textbf{51.9 }                          \\
                                           & DRC-Net (\cite{li2020dual})               & 7.7                            & 17.9                           & 30.5                           \\ \hline
\multicolumn{1}{c|}{\multirow{4}{*}{LoFTR-lite}} & Linear Att. (LoFTR)  (\cite{katharopoulos2020transformers})             & 16.1                           & 32.6                           & 49.0                           \\
\multicolumn{1}{c|}{}                      & PVT ~\citep{wang2021pyramid}        & 16.2                           & 32.7                           & 49.2                           \\
\multicolumn{1}{c|}{}                      & QuadTree-A (ours, $K=8$)        & 16.8                           & 33.4                           & 50.5                           \\
\multicolumn{1}{c|}{}                      & QuadTree-B (ours, $K=8$)        & \textbf{17.4} & \textbf{34.4} & \textbf{51.6} \\ \hline
\multirow{4}{*}{LoFTR}                     & Linear Att. (LoFTR) $\star$ (\cite{sun2021loftr}, 64 GPUs)      & 22.1                           & 40.8                           & 57.6                           \\
                                           & Linear Att. (LoFTR) (\cite{katharopoulos2020transformers})            & 21.1                           & 39.5                           & 56.6                           \\
                                           & QuadTree-B (ours, $K=8$)        & 23.0                           & 41.7                           & 58.5                           \\
                                           & QuadTree-B$\ast$ (ours, $K=16$)    & \textbf{24.9} & \textbf{44.7} & \textbf{61.6}
\end{tabular}
\caption{Results on feature matching. The symbol $\star$ indicates results cited from ~\citep{sun2021loftr}, where the model is trained with a batch size of 64 on 64 GPUs (a more preferable setting than ours). The symbol $\ast$ indicates we use the ViT~\citep{dosovitskiy2020image}-like architecture for transformer blocks. For PVT and our method, we replace the original linear attention in LoFTR with corresponding attentions.}\label{tab:loftr}
\vspace{-7mm}
\end{table}


\textbf{Results.} 
Table~\ref{tab:loftr} shows the AUC of camera pose errors\footnote{Camera pose errors are evaluated as the differences between estimated and ground truth camera orientation and translation direction, both measured in degrees.} under (5$^{\circ}$, 10$^{\circ}$, 20$^{\circ}$). We can see that the SR attention achieves similar results with linear transformer. In comparison, both QuadTree-A and QuadTree-B outperform linear transformer and SR attention by a large margin.  Quadtree-B generally performs better than Quadtree-A. Quadtree-B has 2.6 and 1.9 improvements in terms of AUC@20$^{\circ}$ over linear transformer on LoFTR-lite and LoFTR respectively. To further enhance the results,  we train a model with $K=16$ and leverage a ViT~\citep{dosovitskiy2020image}-like transformer archtecture instead of the original one used in \citep{sun2021loftr}. This model achieves 4 improvements on AUC@20$^{\circ}$ over \citep{sun2021loftr}, where the LoFTR model is trained with a batch size of 64 with 64 GPUs, a more preferable setting leading to slightly better results than our linear transformer implementation shown in Table~\ref{tab:loftr}. 


\subsubsection{Stereo Matching}
Stereo matching aims to find corresponding pixels on epipolar lines between two rectified images. The recent work STTR~\citep{li2020revisiting} applies transformers to feature points between epipolar lines and achieves state-of-the-art performance. Note here, both self- and cross attention are applied along epipolar lines, pixels across different lines are not considered in the attention computation. We replace the standard transformer in STTR~\citep{li2020revisiting} with our quadtree transformer.

\textbf{Setting.} We experiment on the Scene Flow FlyingThings3D \citep{mayer2016large} synthetic dataset, which contains 25,466 images with a resolution of $960 \times 540$.
We build pyramids of four levels to evaluate quadtree attention. 
While the STTR is applied to features of 1/3 of image resolution, we use feature maps of 1/2 of image resolution.
More details about the network are included in Appendix~\ref{sec:stereo_match}.
\begin{table}[]
\centering
\begin{tabular}{l|llll}
                  & EPE (px) & IOU  & Flops (G) & Mem. (MB) \\ \hline
GA-Net~\citep{zhang2019ga}            & 0.89     & /    & /         & /        \\ 
GWC-Net~\citep{guo2019group}           & 0.97     & /    & 305         & 4339        \\ 
Bi3D~\citep{badki2020bi3d}              & 1.16     & /    & 897         & 10031       \\ \hline
STTR (Vanilla Transformer)~\citep{li2020revisiting}              & \textbf{0.45}     & 0.92 & 490       & 8507     \\ 
QuadTree-B (ours, $K=6$) & 0.46     & \textbf{0.99} & \textbf{254 (52\%)}       & \textbf{5381 (63\%)} 
\end{tabular}
\caption{Results of stereo matching. QuadTree-B achieves similar performance as STTR but with significantly lower flops and memory usage.}\label{tab:stereo}
\vspace{-6mm}
\end{table}

\textbf{Results.}
We report EPE (End-Point-Error) in non-occluded regions and IOU (Intersection-over-Union) for occlusion estimation in Table~\ref{tab:stereo} as ~\citep{li2020revisiting}. Computational complexity and memory usage are also reported. 
Compared with STTR based on the standard transformer, our quadtree transformer achieves similar EPE (0.45 px vs 0.46 px) and higher IOU for occlusion estimation, but with much lower computational and memory costs, with only 52\% FLOPs and 63\% memory consumption.

\begin{table}[]
\centering
\small
\begin{tabular}{l|lll}
            & Param (M) & Flops (G) & Top1 (\%)\\ \hline
PVTv2-b0~\citep{wang2021pvtv2}   & 3.7       & 0.6   & 70.5 \\
QuadTree-A-b0 (ours)   & \textbf{3.4}       & 0.6   & 70.9 \\
QuadTree-B-b0 (ours)  & 3.5       & 0.7   & \textbf{72.0} \\ \hline
ResNet18~\citep{he2016deep}    & \textbf{11.7}       & \textbf{1.8}   & 69.8 \\
PVTv1-Tiny~\citep{wang2021pyramid}   & 13.2      & 2.1   & 75.1 \\
PVTv2-b1~\citep{wang2021pvtv2}   & 14.0      & 2.1   & 78.7 \\
QuadTree-B-b1 (ours)  &13.6   & 2.3   &\textbf{80.0}      \\ \hline
ResNet50~\citep{he2016deep}   & 25.1       & 4.1   & 76.4 \\   
ResNeXt50-32x4d~\citep{xie2017aggregated}   & 25.0       & 4.3   & 77.6 \\
RegNetY-4G~\citep{radosavovic2020designing}   & 21.0       & 4.0   & 80.0 \\
DeiT-Small/16~\citep{touvron2021training}   & 22.1       & 4.6   & 79.9 \\
Swin-T~\citep{liu2021swin}    & 29.0        & 4.5   & 81.3 \\
TNT-S~\citep{han2021transformer}& 23.8        & 5.2   & 81.3 \\
CeiT~\citep{yuan2021incorporating} & 24.2        & 4.5   & 82.0 \\
PVTv2-b2~\citep{wang2021pyramid}   & 25.4      & \textbf{4.0}   & 82.0   \\
Focal-T~\citep{yang2021focal}    & 29.1        & 4.9   & 82.2 \\ 
QuadTree-B-b2 (ours) &24.2     & 4.5      &\textbf{82.7}   \\ \hline
ResNet101~\citep{he2016deep}            & 44.7      & 7.9       & 77.4                            \\
ResNeXt101-32x4d~\citep{xie2017aggregated}     & 44.2      & 8.0       & 78.8                            \\
RegNetY-8G~\citep{radosavovic2020designing}           & 39.0      & 8.0       & 81.7                            \\
CvT-21~\citep{wu2021cvt}               & \textbf{32.0}      & 7.1       & 82.5                            \\
PVTv2-b3~\citep{wang2021pyramid}             & 45.2      & \textbf{6.9}       & 83.2                            \\
Quadtree-B-b3 (ours)        & 46.3      & 7.8       & \textbf{83.7}                            \\ \hline
ResNet152~\citep{he2016deep}            & 60.2      & 11.6      & 78.3                            \\
T2T-ViTt-24~\citep{yuan2021tokens}          & 64.0      & 15.0      & 82.2                            \\
Swin-S~\citep{liu2021swin}               & \textbf{50.0}      & \textbf{8.7}       & 83.0                            \\
Focal-Small~\citep{yang2021focal}          & 51.1      & 9.1       & 83.5                            \\
PVTv2-b4~\citep{wang2021pyramid}             & 62.6      & 10.1      & 83.6                            \\
Quadtree-B-b4 (ours)        & 64.2      & 11.5      & \textbf{84.0}                            \\
\end{tabular}
\caption{Image classification results. We report top-1 accuracy on the ImageNet validation set.}\label{tab:imagenet}
\vspace{-3mm}
\end{table}

\subsection{Self-attention task}\label{sec:2dvision}
This section presents results on image classification and object detection. In the past, convolutional neural networks (CNNs) have dominated these tasks for a long time. Recently, vision transformers \citep{dosovitskiy2020image,liu2021swin,wang2021pyramid} show excellent potential on these problems, thanks to their capability in capturing long-range interactions. To compare our method with these vision transformers on image classification, we use the public codes of PVTv2~\citep{wang2021pyramid} and replace all the spatial reduction attention with our quadtree attention. For object detection, we further apply a representative object detection framework, RetinaNet~\citep{lin2017focal}, which is a widely used single-stage object detector.

\subsubsection{Image classification}\label{sec:cls}

\textbf{Settings.} We evaluate image classification on the ImageNet-1K dataset \citep{deng2009imagenet}, which consists of 1.28M training images and 50K validation images from 1,000 categories. We build token pyramids with the coarsest level at a resolution of $7\times 7$ and set $K=8$. We crop and resize the input images to $224\times 224$ pixels and train the model with a mini-batch of 128.  All models are trained for 300 epochs from scratch on 8 GPUs. All the other training settings are the same as in \citep{wang2021pyramid}. We build five different quadtree transformers at different complexity, named as b0, b1, b2, b3, b4. These models are gradually deeper and wider. 
More configuration details can be found in Appendix. \ref{appendix:image_cls}.

\textbf{Results.} We provide the top-1 accuracy of various methods and network settings in Table~\ref{tab:imagenet}. These results are grouped into five sections, each with several methods of similar network complexity, as indicated by the number of parameters. As shown in Table~\ref{tab:imagenet},  QuadTree-B outperforms PVTv2 by 0.4\%-1.5\% in top-1 accuracy with fewer parameters.
Swin Transformer-S adopts local attention and is surpassed by our QuadTree-B-b2 by 1.0\% in top-1 accuracy.
This result proves that global information is important. In general, our quadtree transformer leverages both global information at the coarse level and local information at fine levels, and outperforms both PVTv2 and Swin Transformer.

\begin{table}[]
\centering
\small
\begin{tabular}{l|lllllll}
                & Flops (G) & AP   & AP$_{50}$ & AP$_{75}$ & AP$_S$ &AP$_M$ &AP$_L$  \\ \hline
PVTv2-b0~\citep{wang2021pvtv2}        & 28.3  & 37.2 & 57.2 & 39.5& 23.1 & 40.4 & 49.7 \\

QuadTree-A-b0 (K=32, ours)  & \textbf{16.0}  & 37.0 & 56.8 & 38.9& \textbf{22.8}&
39.7& 50.0 \\
QuadTree-B-b0 (K=32, ours) & 16.5  & \textbf{38.4} & \textbf{58.7} & \textbf{41.1}& 22.5&\textbf{41.7} &51.6 \\ \hline
ResNet18~\citep{he2016deep}   & 38.6  & 31.8 & 49.6 & 33.6 & 16.3 & 34.3 & 43.2 \\
PVTv1-Tiny~\citep{wang2021pyramid}       & 72.5  & 36.7 & 56.9 & 38.9& 22.6 & 38.8 & 50.7 \\
PVTv2-b1~\citep{wang2021pvtv2}       & 78.8  & 41.2 & 61.9 & 43.9& 25.4 & 44.5 & 54.3 \\
Quadtree-B-b1 (K=32, ours) & \textbf{56.2}  &\textbf{42.6}  &\textbf{63.6}  &\textbf{45.3}& \textbf{26.8}&
\textbf{46.1}& \textbf{57.2}    \\ \hline
ResNet50~\citep{he2016deep}       &  87.3 & 36.3 & 55.3 & 38.6& 19.3 & 40.0 & 48.8 \\
ResNet101~\citep{he2016deep}       &  166.3 & 38.5 &  57.8 & 41.2&  21.4 &  42.6 &  51.1 \\
ResNeXt101-32x4d~\citep{xie2017aggregated}       & 170.2 & 39.9 & 59.6 &  42.7& 22.3 &  44.2 & 52.5 \\
PVTv1-small~\citep{wang2021pyramid}       & 139.8  & 36.7 & 56.9 & 38.9& 25.0 & 42.9 &  55.7 \\
PVTv2-b2~\citep{wang2021pyramid}     & 149.1 &44.6 &65.6  &47.6& 27.4 & 48.8 & 58.6 \\
QuadTree-B-b2 (K=32, ours)& \textbf{108.6}  & \textbf{46.2}& \textbf{67.2} &\textbf{49.5}& \textbf{29.0} &\textbf{50.1} &\textbf{61.8}
\\ \hline
PVTv1-Medium~\citep{wang2021pyramid}                                                 &      237.4                           & 41.9                           & 63.1                           & 44.3                           & 25.0                           & 44.9                           & 57.6                           \\
PVTv2-b3~\citep{wang2021pvtv2}                                                     &      243.0                           & 45.9                           & 66.8                           & 49.3                           & 28.6                           & 49.8                           & 61.4                           \\
QuadTree-B-b3 (ours)                                              &          \textbf{193.9}                         & \textbf{47.3}       & \textbf{68.2}       & \textbf{50.6}       & \textbf{30.4}       & \textbf{51.3}       & \textbf{62.9}       \\ \hline
PVTv1-Large~\citep{wang2021pyramid}                                                  &     346.6                            & 42.6                           & 63.7                           & 45.4                           & 25.8                           & 46.0                           & 58.4                           \\
PVTv2-b4~\citep{wang2021pvtv2}                                                     &      353.3                           & 46.1                           & 66.9                           & 49.2                           & 28.4                           & 50.0                           & 62.2                           \\
QuadTree-B-b4 (ours)                                              &      \textbf{283.9}                           & \textbf{47.9}       & \textbf{69.1}       & \textbf{51.3}       & \textbf{29.4}       & \textbf{52.2}       & \textbf{63.9}      

\end{tabular}
\caption{Object detection results on COCO val2017 with RetinaNet. We use PVTv2 backbone and replace the reduction attention with quadtree attention. `Flops' is the backbone flops for input image size of $800\times 1,333$.}
\label{tab:detection}
\vspace{-3mm}
\end{table}

\subsubsection{Object detection}

\textbf{Settings.} We experiment on the COCO dataset. All models are trained on COCO train 2017 (118k images) and evaluated on val 2017 (5k images). We initialize the quadtree backbone with the weights pre-trained on ImageNet. We adopt the same setting as PVTv2, training the model with a batch size of 16 and AdamW optimizer with an initial learning rate of $1\times 10^{-4}$ for 12 epochs. We use the standard metric average precision to evaluate our method.

\textbf{Results.} We mainly compare our method with PVTv2, ResNet \citep{he2016deep}, and ResNeXt \citep{xie2017aggregated} using detection framework of RetinaNet~\citep{lin2017focal}, which are state-of-the-art backbones for dense prediction. Table~\ref{tab:detection} lists the average precision of different methods and their backbone flops for images of resolution of $800\times 1,333$. Benefiting from the coarse to fine mechanism, a small $K$ is enough for our method. Thus, the computation can be reduced when using high resolution images.
We can see that QuadTree-B achieves higher performance, but with much fewer flops than PVTv2. 
Our quadtree transformer also outperforms ResNet and ResNeXt. For example, QuadTree-B-b2 outperform ResNet101 and ResNeXt101-32x4d by 7.7 AP and 6.3 AP respectively with about 40\% backbone flops reduction. We also show Mask-RCNN results~\citep{he2017mask} in Appendix. \ref{apx:experiments}.

\subsection{Comparison with other attention mechanisms}\label{sec:analysis}


For a fair comparison with other attention mechanisms, we test these attention mechanisms under the same backbone and training settings. Specifically, we replace the original attention module in PVTv2-b0 with the attention method used in Swin Transformer and Focal Transformer. For more fair comparison, we adopt the same positional encoding LePE ~\citep{dong2021cswin} to PVTv2, Swin and Focal transformer. As shown in Table \ref{tab:attn_compare}, QuadTree attention obtain consistently better performance than Swin and PVTv2 in both classification task and detection task. Compared with focal attention, our method gets 0.9 higher AP in object detection, which might be because that QuadTree attention can always cover the whole images, while Focal attention only covers $1/6$ of the image in the first stage. More experiments on Swin-like architecture can be found in Appendix \ref{apx:experiments}. 

For cross attention tasks, we also provide visualization of attention score as shown in Fig.\ref{fig:score_vis} in Appendix~\ref{apx:experiments}. Our method can attend to much more related regions than PVT ~\citep{wang2021pvtv2} and Linear attention ~\citep{katharopoulos2020transformers}.

\begin{table}[]
\centering
\small
\begin{tabular}{l|ll|llll}
                                                             & \multicolumn{2}{l|}{ImageNet-1K}                               & \multicolumn{4}{l}{COCO (RetinaNet)}                                                                                             \\ \hline
                                                             & Flops (G)                     & Top-1 (\%)                     & Mem.  (MB)                    & AP                             & AP$_{50}$                      & AP$_{75}$                      \\ \hline
PVTv2 ~\citep{wang2021pvtv2}           & \textbf{0.6} & 70.5                           & 574                           & 37.2                           & 57.2                           & 39.5                           \\
PVTv2+LePE ~\citep{dong2021cswin}      & \textbf{0.6} & 70.9                           & 574                           & 37.6                           & 57.8                           & 39.9                           \\ \hline
Swin ~\citep{liu2021swin}              & \textbf{0.6} & 70.5                           & \textbf{308} & 35.3                           & 54.2                           & 37.4                           \\
Swin+LePE                                                    & 0.6                           & 70.7                           & \textbf{308}                           & 35.8                           & 55.3                           & 37.7                           \\ \hline
Focal Attention ~\citep{yang2021focal} & 0.7                           & 71.6                           & 732                           & 37.5                           & 57.6                           & 39.5                           \\
Focal Attention+LePE                                         & 0.7                           & 71.5                           & 732                           & 37.1                           & 57.0                           & 39.4                           \\ \hline
QuadTree-B                                                   & \textbf{0.6} & \textbf{72.0} & 339                           & \textbf{38.4} & \textbf{58.8} & \textbf{41.1}
\end{tabular}
\caption{To fairly compare with Swin, PVT, Focal attention and our method, we replace the attention module in PVTv2-b0 with different types of attention and same position encoding method LePE and run image classification and object detection respectively.}\label{tab:attn_compare}
\vspace{-7mm}
\end{table}

\section{Conclusion}
We introduce QuadTree Attention to reduce the computational complexity of vision transformers from quadratic to linear. Quadtree transformers build token pyramids and compute attention in a coarse-to-fine manner. At each level, top $K$ regions with the highest attention scores are selected, such that in finer level, computation in irrelevant regions can be quickly skipped. Quadtree attention can be applied to cross attention as well as self-attention. It achieves state-of-the-art performance in various tasks including feature matching, stereo, image classification, and object detection.  

\bibliography{egbib}

\begin{thebibliography}{46}
\providecommand{\natexlab}[1]{#1}
\providecommand{\url}[1]{\texttt{#1}}
\expandafter\ifx\csname urlstyle\endcsname\relax
  \providecommand{\doi}[1]{doi: #1}\else
  \providecommand{\doi}{doi: \begingroup \urlstyle{rm}\Url}\fi

\bibitem[Badki et~al.(2020)Badki, Troccoli, Kim, Kautz, Sen, and
  Gallo]{badki2020bi3d}
Abhishek Badki, Alejandro Troccoli, Kihwan Kim, Jan Kautz, Pradeep Sen, and
  Orazio Gallo.
\newblock Bi3d: Stereo depth estimation via binary classifications.
\newblock In \emph{Proceedings of the IEEE/CVF Conference on Computer Vision
  and Pattern Recognition}, pp.\  1600--1608, 2020.

\bibitem[Beltagy et~al.(2020)Beltagy, Peters, and Cohan]{beltagy2020longformer}
Iz~Beltagy, Matthew~E Peters, and Arman Cohan.
\newblock Longformer: The long-document transformer.
\newblock \emph{arXiv preprint arXiv:2004.05150}, 2020.

\bibitem[Chen et~al.(2021)Chen, Luo, Zhang, Zhou, Bai, Hu, Tai, and
  Quan]{chen2021learning}
Hongkai Chen, Zixin Luo, Jiahui Zhang, Lei Zhou, Xuyang Bai, Zeyu Hu, Chiew-Lan
  Tai, and Long Quan.
\newblock Learning to match features with seeded graph matching network.
\newblock \emph{arXiv preprint arXiv:2108.08771}, 2021.

\bibitem[Choromanski et~al.(2020)Choromanski, Likhosherstov, Dohan, Song, Gane,
  Sarlos, Hawkins, Davis, Mohiuddin, Kaiser, et~al.]{choromanski2020rethinking}
Krzysztof Choromanski, Valerii Likhosherstov, David Dohan, Xingyou Song,
  Andreea Gane, Tamas Sarlos, Peter Hawkins, Jared Davis, Afroz Mohiuddin,
  Lukasz Kaiser, et~al.
\newblock Rethinking attention with performers.
\newblock \emph{arXiv preprint arXiv:2009.14794}, 2020.

\bibitem[Cuturi(2013)]{cuturi2013sinkhorn}
Marco Cuturi.
\newblock Sinkhorn distances: Lightspeed computation of optimal transport.
\newblock \emph{Advances in neural information processing systems},
  26:\penalty0 2292--2300, 2013.

\bibitem[Dai et~al.(2017)Dai, Chang, Savva, Halber, Funkhouser, and
  Nie{\ss}ner]{dai2017scannet}
Angela Dai, Angel~X Chang, Manolis Savva, Maciej Halber, Thomas Funkhouser, and
  Matthias Nie{\ss}ner.
\newblock Scannet: Richly-annotated 3d reconstructions of indoor scenes.
\newblock In \emph{Proceedings of the IEEE conference on computer vision and
  pattern recognition}, pp.\  5828--5839, 2017.

\bibitem[Deng et~al.(2009)Deng, Dong, Socher, Li, Li, and
  Fei-Fei]{deng2009imagenet}
Jia Deng, Wei Dong, Richard Socher, Li-Jia Li, Kai Li, and Li~Fei-Fei.
\newblock Imagenet: A large-scale hierarchical image database.
\newblock In \emph{2009 IEEE conference on computer vision and pattern
  recognition}, pp.\  248--255. Ieee, 2009.

\bibitem[DeTone et~al.(2018)DeTone, Malisiewicz, and
  Rabinovich]{detone2018superpoint}
Daniel DeTone, Tomasz Malisiewicz, and Andrew Rabinovich.
\newblock Superpoint: Self-supervised interest point detection and description.
\newblock In \emph{Proceedings of the IEEE conference on computer vision and
  pattern recognition workshops}, pp.\  224--236, 2018.

\bibitem[Dong et~al.(2021)Dong, Bao, Chen, Zhang, Yu, Yuan, Chen, and
  Guo]{dong2021cswin}
Xiaoyi Dong, Jianmin Bao, Dongdong Chen, Weiming Zhang, Nenghai Yu, Lu~Yuan,
  Dong Chen, and Baining Guo.
\newblock Cswin transformer: A general vision transformer backbone with
  cross-shaped windows.
\newblock \emph{arXiv preprint arXiv:2107.00652}, 2021.

\bibitem[Dosovitskiy et~al.(2020)Dosovitskiy, Beyer, Kolesnikov, Weissenborn,
  Zhai, Unterthiner, Dehghani, Minderer, Heigold, Gelly,
  et~al.]{dosovitskiy2020image}
Alexey Dosovitskiy, Lucas Beyer, Alexander Kolesnikov, Dirk Weissenborn,
  Xiaohua Zhai, Thomas Unterthiner, Mostafa Dehghani, Matthias Minderer, Georg
  Heigold, Sylvain Gelly, et~al.
\newblock An image is worth 16x16 words: Transformers for image recognition at
  scale.
\newblock \emph{arXiv preprint arXiv:2010.11929}, 2020.

\bibitem[Germain et~al.(2021)Germain, Lepetit, and Bourmaud]{germain2021visual}
Hugo Germain, Vincent Lepetit, and Guillaume Bourmaud.
\newblock Visual correspondence hallucination: Towards geometric reasoning.
\newblock \emph{arXiv preprint arXiv:2106.09711}, 2021.

\bibitem[Guo et~al.(2019)Guo, Yang, Yang, Wang, and Li]{guo2019group}
Xiaoyang Guo, Kai Yang, Wukui Yang, Xiaogang Wang, and Hongsheng Li.
\newblock Group-wise correlation stereo network.
\newblock In \emph{Proceedings of the IEEE/CVF Conference on Computer Vision
  and Pattern Recognition}, pp.\  3273--3282, 2019.

\bibitem[Han et~al.(2021)Han, Xiao, Wu, Guo, Xu, and Wang]{han2021transformer}
Kai Han, An~Xiao, Enhua Wu, Jianyuan Guo, Chunjing Xu, and Yunhe Wang.
\newblock Transformer in transformer.
\newblock \emph{arXiv preprint arXiv:2103.00112}, 2021.

\bibitem[He et~al.(2016)He, Zhang, Ren, and Sun]{he2016deep}
Kaiming He, Xiangyu Zhang, Shaoqing Ren, and Jian Sun.
\newblock Deep residual learning for image recognition.
\newblock In \emph{Proceedings of the IEEE conference on computer vision and
  pattern recognition}, pp.\  770--778, 2016.

\bibitem[He et~al.(2017)He, Gkioxari, Doll{\'a}r, and Girshick]{he2017mask}
Kaiming He, Georgia Gkioxari, Piotr Doll{\'a}r, and Ross Girshick.
\newblock Mask r-cnn.
\newblock In \emph{Proceedings of the IEEE international conference on computer
  vision}, pp.\  2961--2969, 2017.

\bibitem[Katharopoulos et~al.(2020)Katharopoulos, Vyas, Pappas, and
  Fleuret]{katharopoulos2020transformers}
Angelos Katharopoulos, Apoorv Vyas, Nikolaos Pappas, and Fran{\c{c}}ois
  Fleuret.
\newblock Transformers are rnns: Fast autoregressive transformers with linear
  attention.
\newblock In \emph{International Conference on Machine Learning}, pp.\
  5156--5165. PMLR, 2020.

\bibitem[Lee et~al.(2019)Lee, Lee, Kim, Kosiorek, Choi, and Teh]{lee2019set}
Juho Lee, Yoonho Lee, Jungtaek Kim, Adam Kosiorek, Seungjin Choi, and Yee~Whye
  Teh.
\newblock Set transformer: A framework for attention-based
  permutation-invariant neural networks.
\newblock In \emph{International Conference on Machine Learning}, pp.\
  3744--3753. PMLR, 2019.

\bibitem[Li et~al.(2020)Li, Han, Li, and Prisacariu]{li2020dual}
Xinghui Li, Kai Han, Shuda Li, and Victor Prisacariu.
\newblock Dual-resolution correspondence networks.
\newblock \emph{Advances in Neural Information Processing Systems}, 33, 2020.

\bibitem[Li et~al.(2021)Li, Liu, Drenkow, Ding, Creighton, Taylor, and
  Unberath]{li2020revisiting}
Zhaoshuo Li, Xingtong Liu, Nathan Drenkow, Andy Ding, Francis~X Creighton,
  Russell~H Taylor, and Mathias Unberath.
\newblock Revisiting stereo depth estimation from a sequence-to-sequence
  perspective with transformers.
\newblock \emph{IEEE/CVF International Conference on Computer Vision}, 2021.

\bibitem[Li \& Snavely(2018)Li and Snavely]{li2018megadepth}
Zhengqi Li and Noah Snavely.
\newblock Megadepth: Learning single-view depth prediction from internet
  photos.
\newblock In \emph{Proceedings of the IEEE Conference on Computer Vision and
  Pattern Recognition}, pp.\  2041--2050, 2018.

\bibitem[Lin et~al.(2014)Lin, Maire, Belongie, Hays, Perona, Ramanan,
  Doll{\'a}r, and Zitnick]{lin2014microsoft}
Tsung-Yi Lin, Michael Maire, Serge Belongie, James Hays, Pietro Perona, Deva
  Ramanan, Piotr Doll{\'a}r, and C~Lawrence Zitnick.
\newblock Microsoft coco: Common objects in context.
\newblock In \emph{European conference on computer vision}, pp.\  740--755.
  Springer, 2014.

\bibitem[Lin et~al.(2017)Lin, Goyal, Girshick, He, and
  Doll{\'a}r]{lin2017focal}
Tsung-Yi Lin, Priya Goyal, Ross Girshick, Kaiming He, and Piotr Doll{\'a}r.
\newblock Focal loss for dense object detection.
\newblock In \emph{Proceedings of the IEEE international conference on computer
  vision}, pp.\  2980--2988, 2017.

\bibitem[Liu et~al.(2021)Liu, Lin, Cao, Hu, Wei, Zhang, Lin, and
  Guo]{liu2021swin}
Ze~Liu, Yutong Lin, Yue Cao, Han Hu, Yixuan Wei, Zheng Zhang, Stephen Lin, and
  Baining Guo.
\newblock Swin transformer: Hierarchical vision transformer using shifted
  windows.
\newblock \emph{arXiv preprint arXiv:2103.14030}, 2021.

\bibitem[Luo et~al.(2019)Luo, Shen, Zhou, Zhang, Yao, Li, Fang, and
  Quan]{luo2019contextdesc}
Zixin Luo, Tianwei Shen, Lei Zhou, Jiahui Zhang, Yao Yao, Shiwei Li, Tian Fang,
  and Long Quan.
\newblock Contextdesc: Local descriptor augmentation with cross-modality
  context.
\newblock In \emph{Proceedings of the IEEE/CVF Conference on Computer Vision
  and Pattern Recognition}, pp.\  2527--2536, 2019.

\bibitem[Mayer et~al.(2016)Mayer, Ilg, Hausser, Fischer, Cremers, Dosovitskiy,
  and Brox]{mayer2016large}
Nikolaus Mayer, Eddy Ilg, Philip Hausser, Philipp Fischer, Daniel Cremers,
  Alexey Dosovitskiy, and Thomas Brox.
\newblock A large dataset to train convolutional networks for disparity,
  optical flow, and scene flow estimation.
\newblock In \emph{Proceedings of the IEEE conference on computer vision and
  pattern recognition}, pp.\  4040--4048, 2016.

\bibitem[Radosavovic et~al.(2020)Radosavovic, Kosaraju, Girshick, He, and
  Doll{\'a}r]{radosavovic2020designing}
Ilija Radosavovic, Raj~Prateek Kosaraju, Ross Girshick, Kaiming He, and Piotr
  Doll{\'a}r.
\newblock Designing network design spaces.
\newblock In \emph{Proceedings of the IEEE/CVF Conference on Computer Vision
  and Pattern Recognition}, pp.\  10428--10436, 2020.

\bibitem[Ranftl et~al.(2021)Ranftl, Bochkovskiy, and Koltun]{ranftl2021vision}
Ren{\'e} Ranftl, Alexey Bochkovskiy, and Vladlen Koltun.
\newblock Vision transformers for dense prediction.
\newblock \emph{arXiv preprint arXiv:2103.13413}, 2021.

\bibitem[Sarlin et~al.(2020)Sarlin, DeTone, Malisiewicz, and
  Rabinovich]{sarlin2020superglue}
Paul-Edouard Sarlin, Daniel DeTone, Tomasz Malisiewicz, and Andrew Rabinovich.
\newblock Superglue: Learning feature matching with graph neural networks.
\newblock In \emph{Proceedings of the IEEE/CVF conference on computer vision
  and pattern recognition}, pp.\  4938--4947, 2020.

\bibitem[Shao et~al.(2020)Shao, Tang, Pan, Tan, Wang, and Luo]{shao2020channel}
Wenqi Shao, Shitao Tang, Xingang Pan, Ping Tan, Xiaogang Wang, and Ping Luo.
\newblock Channel equilibrium networks for learning deep representation.
\newblock In \emph{International Conference on Machine Learning}, pp.\
  8645--8654. PMLR, 2020.

\bibitem[Shao et~al.(2021)Shao, Ge, Zhang, Xu, Wang, Shan, and
  Luo]{shao2021dynamic}
Wenqi Shao, Yixiao Ge, Zhaoyang Zhang, Xuyuan Xu, Xiaogang Wang, Ying Shan, and
  Ping Luo.
\newblock Dynamic token normalization improves vision transformer.
\newblock \emph{arXiv preprint arXiv:2112.02624}, 2021.

\bibitem[Sun et~al.(2021)Sun, Shen, Wang, Bao, and Zhou]{sun2021loftr}
Jiaming Sun, Zehong Shen, Yuang Wang, Hujun Bao, and Xiaowei Zhou.
\newblock Loftr: Detector-free local feature matching with transformers.
\newblock In \emph{Proceedings of the IEEE/CVF Conference on Computer Vision
  and Pattern Recognition}, pp.\  8922--8931, 2021.

\bibitem[Touvron et~al.(2021)Touvron, Cord, Douze, Massa, Sablayrolles, and
  J{\'e}gou]{touvron2021training}
Hugo Touvron, Matthieu Cord, Matthijs Douze, Francisco Massa, Alexandre
  Sablayrolles, and Herv{\'e} J{\'e}gou.
\newblock Training data-efficient image transformers \& distillation through
  attention.
\newblock In \emph{International Conference on Machine Learning}, pp.\
  10347--10357. PMLR, 2021.

\bibitem[Wang et~al.(2021{\natexlab{a}})Wang, Wang, Wang, Lin, Chang, Xie, Li,
  and Jin]{wang2021kvt}
Pichao Wang, Xue Wang, Fan Wang, Ming Lin, Shuning Chang, Wen Xie, Hao Li, and
  Rong Jin.
\newblock Kvt: k-nn attention for boosting vision transformers.
\newblock \emph{arXiv preprint arXiv:2106.00515}, 2021{\natexlab{a}}.

\bibitem[Wang et~al.(2020)Wang, Li, Khabsa, Fang, and Ma]{wang2020linformer}
Sinong Wang, Belinda~Z Li, Madian Khabsa, Han Fang, and Hao Ma.
\newblock Linformer: Self-attention with linear complexity.
\newblock \emph{arXiv preprint arXiv:2006.04768}, 2020.

\bibitem[Wang et~al.(2021{\natexlab{b}})Wang, Xie, Li, Fan, Song, Liang, Lu,
  Luo, and Shao]{wang2021pvtv2}
Wenhai Wang, Enze Xie, Xiang Li, Deng-Ping Fan, Kaitao Song, Ding Liang, Tong
  Lu, Ping Luo, and Ling Shao.
\newblock Pvtv2: Improved baselines with pyramid vision transformer.
\newblock \emph{arXiv preprint arXiv:2106.13797}, 2021{\natexlab{b}}.

\bibitem[Wang et~al.(2021{\natexlab{c}})Wang, Xie, Li, Fan, Song, Liang, Lu,
  Luo, and Shao]{wang2021pyramid}
Wenhai Wang, Enze Xie, Xiang Li, Deng-Ping Fan, Kaitao Song, Ding Liang, Tong
  Lu, Ping Luo, and Ling Shao.
\newblock Pyramid vision transformer: A versatile backbone for dense prediction
  without convolutions.
\newblock \emph{arXiv preprint arXiv:2102.12122}, 2021{\natexlab{c}}.

\bibitem[Wu et~al.(2021)Wu, Xiao, Codella, Liu, Dai, Yuan, and
  Zhang]{wu2021cvt}
Haiping Wu, Bin Xiao, Noel Codella, Mengchen Liu, Xiyang Dai, Lu~Yuan, and Lei
  Zhang.
\newblock Cvt: Introducing convolutions to vision transformers.
\newblock \emph{arXiv preprint arXiv:2103.15808}, 2021.

\bibitem[Xie et~al.(2017)Xie, Girshick, Doll{\'a}r, Tu, and
  He]{xie2017aggregated}
Saining Xie, Ross Girshick, Piotr Doll{\'a}r, Zhuowen Tu, and Kaiming He.
\newblock Aggregated residual transformations for deep neural networks.
\newblock In \emph{Proceedings of the IEEE conference on computer vision and
  pattern recognition}, pp.\  1492--1500, 2017.

\bibitem[Yang et~al.(2021)Yang, Li, Zhang, Dai, Xiao, Yuan, and
  Gao]{yang2021focal}
Jianwei Yang, Chunyuan Li, Pengchuan Zhang, Xiyang Dai, Bin Xiao, Lu~Yuan, and
  Jianfeng Gao.
\newblock Focal self-attention for local-global interactions in vision
  transformers.
\newblock \emph{arXiv preprint arXiv:2107.00641}, 2021.

\bibitem[Ying et~al.(2018)Ying, You, Morris, Ren, Hamilton, and
  Leskovec]{ying2018hierarchical}
Rex Ying, Jiaxuan You, Christopher Morris, Xiang Ren, William~L Hamilton, and
  Jure Leskovec.
\newblock Hierarchical graph representation learning with differentiable
  pooling.
\newblock \emph{arXiv preprint arXiv:1806.08804}, 2018.

\bibitem[Yuan et~al.(2021{\natexlab{a}})Yuan, Guo, Liu, Zhou, Yu, and
  Wu]{yuan2021incorporating}
Kun Yuan, Shaopeng Guo, Ziwei Liu, Aojun Zhou, Fengwei Yu, and Wei Wu.
\newblock Incorporating convolution designs into visual transformers.
\newblock \emph{arXiv preprint arXiv:2103.11816}, 2021{\natexlab{a}}.

\bibitem[Yuan et~al.(2021{\natexlab{b}})Yuan, Chen, Wang, Yu, Shi, Jiang, Tay,
  Feng, and Yan]{yuan2021tokens}
Li~Yuan, Yunpeng Chen, Tao Wang, Weihao Yu, Yujun Shi, Zihang Jiang, Francis~EH
  Tay, Jiashi Feng, and Shuicheng Yan.
\newblock Tokens-to-token vit: Training vision transformers from scratch on
  imagenet.
\newblock \emph{arXiv preprint arXiv:2101.11986}, 2021{\natexlab{b}}.

\bibitem[Zaheer et~al.(2020)Zaheer, Guruganesh, Dubey, Ainslie, Alberti,
  Ontanon, Pham, Ravula, Wang, Yang, et~al.]{zaheer2020big}
Manzil Zaheer, Guru Guruganesh, Kumar~Avinava Dubey, Joshua Ainslie, Chris
  Alberti, Santiago Ontanon, Philip Pham, Anirudh Ravula, Qifan Wang, Li~Yang,
  et~al.
\newblock Big bird: Transformers for longer sequences.
\newblock In \emph{NeurIPS}, 2020.

\bibitem[Zhang et~al.(2019{\natexlab{a}})Zhang, Prisacariu, Yang, and
  Torr]{zhang2019ga}
Feihu Zhang, Victor Prisacariu, Ruigang Yang, and Philip~HS Torr.
\newblock Ga-net: Guided aggregation net for end-to-end stereo matching.
\newblock In \emph{Proceedings of the IEEE/CVF Conference on Computer Vision
  and Pattern Recognition}, pp.\  185--194, 2019{\natexlab{a}}.

\bibitem[Zhang et~al.(2019{\natexlab{b}})Zhang, Sun, Luo, Yao, Zhou, Shen,
  Chen, Quan, and Liao]{zhang2019learning}
Jiahui Zhang, Dawei Sun, Zixin Luo, Anbang Yao, Lei Zhou, Tianwei Shen, Yurong
  Chen, Long Quan, and Hongen Liao.
\newblock Learning two-view correspondences and geometry using order-aware
  network.
\newblock In \emph{Proceedings of the IEEE/CVF International Conference on
  Computer Vision}, pp.\  5845--5854, 2019{\natexlab{b}}.

\bibitem[Zhao et~al.(2019)Zhao, Lin, Zhang, Ren, Su, and Sun]{zhao2019explicit}
Guangxiang Zhao, Junyang Lin, Zhiyuan Zhang, Xuancheng Ren, Qi~Su, and Xu~Sun.
\newblock Explicit sparse transformer: Concentrated attention through explicit
  selection.
\newblock \emph{arXiv preprint arXiv:1912.11637}, 2019.

\end{thebibliography}
\bibliographystyle{iclr2022_conference}

\appendix
\section{Appendix}
\subsection{Complexity analysis}\label{apx:complexity}
In this section, we analyze the computational complexity of quadtree attention. Suppose the lengths of the query tokens, key tokens, and value tokens are all $H\times W$. We build token pyramids of $L$ levels, the $l^{th}$ level has a token length of $\frac{H W}{4^{l-1}}$. 
The flops of computing quadtree attention is,
\begin{align*}
     \text{Flops}&=2(H_0^2W_0^2 + \sum_{l-2}^{L-1}\frac{4KHW}{4^{l-1}}) \\
     &=2(H_0^2W_0^2 + \frac{4}{3}(1-4^{1-L})KHW).
\end{align*}
Here, $H_0$ and $W_0$ are the height and width of the coarsest level of token pyramids. Therefore, $H_0^2W_0^2$ is a constant and the computational complexity is $O(KHW)$.
Since $K$ is a constant number, the complexity of quadtree attention is linear to the number of tokens.

\section{Additional experiments and implementation details}\label{sec:task_intro}

\begin{table}[]
\centering
\begin{tabular}{l|lll}
                        & AUC@5$^{\circ}$ & AUC@10$^{\circ}$ & AUC@20$^{\circ}$ \\ \hline
DRC-Net~\citep{li2020dual}                 & 27.0 & 43.0  & 58.3  \\
SuperPoint + SuperGlue~\citep{sarlin2020superglue}  & 42.2 & 61.2  & 76.0  \\
LoFTR~\citep{sun2021loftr}                   & 52.8 & 69.2  & 81.2  \\
QuadTree-B (ours, K=16) & \textbf{54.6} & \textbf{70.5}  & \textbf{82.2}
\end{tabular}
\caption{Feature matching results on megadepth. Our method obtains better performance than other methods.}\label{tab:megadepth}
\end{table}

\subsection{Feature Matching}\label{appendix:image match}

\textbf{Implementation details.} We train and evaluate the model in ScanNet~\citep{dai2017scannet}, where 230M image pairs is sampled for training, with overlapping scores between 0.4 and 0.8. ScanNet provides RGB images, depth maps, and ground truth camera poses on a well-defined training and testing split. Following the same evaluation settings as \cite{sarlin2020superglue} and \cite{sun2021loftr}, we evaluate our method on the 1,500 testing pairs from ~\citep{sarlin2020superglue}. For both trainig and testing, all images and depth maps are resized to 640 × 480. Following ~\citep{sun2021loftr}, we compute the camera pose by solving the essential matrix from predicted matches with RANSAC. We report the AUC of the pose error at thresholds (5$^{\circ}$, 10$^{\circ}$, 20$^{\circ}$), where the pose error is defined as the maximum of angular error in rotation and translation. We only replace the coarse level transformer with quadtree attention.

\textbf{Results of megadepth.} We show our results on Megadepth~\citep{li2018megadepth} in Table \ref{tab:megadepth}. We can see our method outperforms others by a large margin.

\subsection{Stereo matching}\label{sec:stereo_match}
Our network is based on the STTR \citep{li2020revisiting}, where we replace the standard transformer with our quadtree transformer. 
The network consists of a CNN backbone which outputs feature maps of 1/2 image resolution, a quadtree transformer with both self- and cross attention, a regression head with optimal transport layers ~\citep{cuturi2013sinkhorn}, and a context adjust layer to refine the disparity. Six self- and cross attention layers are used with 128 channels. We build pyramids with four levels for quadtree attention, and apply the Sinkhorn algorithm ~\citep{cuturi2013sinkhorn} for 10 iteration for optimal transport. 
We follow STTR to train the network, with 15 epochs of AdamW optimizer. OneCycle learning rate scheduler is used with a leaning rate of 6e-4 and a batch size of 8.

\begin{table}[]
\centering
\small
\begin{tabular}{l|llllll}
                                                             & AP$^b$   & AP$_{50}^b$ & AP$_{75}^b$ & AP$^m$ & AP$^m_{50}$ & AP$^m_{75}$ \\ \hline
PVTv2-b0~\citep{wang2021pvtv2}                                                     & 38.2 & 60.5      & 40.7      & 36.2   & 57.8   & 38.6   \\
QuadTree-B-b0 (K=32, ours)                                   & \textbf{38.8} & \textbf{60.7}      & \textbf{42.1}      & \textbf{36.5}   & \textbf{58.0}   & \textbf{39.1}   \\ \hline
ResNet18~\citep{he2016deep}                & 34.0 & 54.0      & 36.7      & 31.2   & 51.0   & 32.7   \\
PVTv1-Tiny~\citep{wang2021pyramid}         & 36.7 & 59.2      & 39.3      & 35.1   & 56.7   & 37.3   \\
PVTv2-b1~\citep{wang2021pvtv2}             & 41.8 & 64.3      & 45.9      & 38.8   & 61.2   & 41.6   \\
Quadtree-B-b1 (K=32, ours)                                   & \textbf{43.5} & \textbf{65.6}      & \textbf{47.6}      & \textbf{40.1}   & \textbf{62.6}   & \textbf{43.3}   \\ \hline
ResNet50~\citep{he2016deep}                & 38.0 & 58.6      & 41.4      & 34.4   & 55.1   & 36.7   \\
ResNet101~\citep{he2016deep}              & 40.4 & 61.1      & 44.2      & 36.4   & 57.7   & 38.8   \\
ResNeXt101-32x4d~\citep{xie2017aggregated} & 41.9 & 62.5      & 45.9      & 37.5   & 59.4   & 40.2   \\
PVTv1-small~\citep{wang2021pyramid}        & 40.4 & 62.9      & 43.8      & 37.8   & 60.1   & 40.3   \\
PVTv2-b2~\citep{wang2021pvtv2}           & 45.3 & 67.1      & 49.6      & 41.2   & 64.2   & 44.4   \\
QuadTree-B-b2 (K=32, ours)                                   & \textbf{46.7} & \textbf{68.5}      & \textbf{51.2}      & \textbf{42.4}   & \textbf{65.7}   & \textbf{45.7}   \\ \hline
PVTv1-Medium~\citep{wang2021pyramid}                                                 & 42.0 & 64.4      & 45.6      & 39.0   & 61.6   & 42.1   \\
PVTv2-b3~\citep{wang2021pvtv2}                                                     & 45.9 & 66.8      & 49.3      & 28.6   & 49.8   & 61.4   \\
QuadTree-B-b3                                                & \textbf{48.3} & \textbf{69.6}      & \textbf{52.8}      & \textbf{43.3}   & \textbf{66.8}   & \textbf{46.6}   \\ \hline
PVTv1-Large~\citep{wang2021pyramid}                                                  & 42.9 & 65.0      & 46.6      & 39.5   & 61.9   & 42.5   \\
PVTv2-b4~\citep{wang2021pvtv2}                                                      & 47.5 & 68.7      & 52.0      & 42.7   & 66.1   & 46.1   \\
QuadTree-B-b4                                                & \textbf{48.6} & \textbf{69.5}      & \textbf{53.3}      & \textbf{43.6}   & \textbf{66.9}   & \textbf{47.4}
\end{tabular}
\caption{Object detection results on COCO val2017 with Mask-RCNN. We use PVTv2 backbone and replace the reduction attention with quadtree attention.}\label{tab:maskrcnn1x}
\end{table}

\subsection{Image classification}\label{appendix:image_cls}
This paragraph introduces the details of PVTv2-b0, b1, b2, b3, b4. All these five networks have 4 stages. Each stage is down-sampled from the previous stage by a stride of 2. The feature resolutions for each stage are $\frac{H}{4}\times \frac{W}{4}$, $\frac{H}{8}\times \frac{W}{8}$, $\frac{H}{16}\times \frac{W}{16}$ and $\frac{H}{32}\times \frac{W}{32}$ respectively, where $H$ and $W$ is the image height and width. For each stage, $M$ quadtree transformers are used with a channel number of $I$ and head number of $J$. For the network PVTv2-b0, the parameters $M$, $I$, $J$ are set to $[2,2,2,2]$, $[32, 64, 160, 256]$, $[1, 2, 5, 8]$ at each stage respectively. For the network PVTv2-b1, the parameters $M$, $I$, $J$ are set to $[2,2,2,2]$, $[64, 128, 320, 512]$, $[1, 2, 5, 8]$ respectively. For PVTv2-b2, the parameters $M$, $I$, $J$ are set to $[3,4,6,3]$, $[64, 128, 320, 512]$, $[1, 2, 5, 8]$ respectively. For PVTv2-b3, the parameters $M$, $I$, $J$ are set to $[3, 4, 18, 3]$, $[64, 128, 320, 512]$, $[1, 2, 5, 8]$ respectively. For PVTv2-b4, the parameters $M$, $I$, $J$ are set to $[3, 8, 27, 3]$, $[64, 128, 320, 512]$, $[1, 2, 5, 8]$ respectively.

\subsection{Object detection and instance segmentation}
We show the object detection and instance segmentation results of Mask-RCNN~\citep{he2017mask} in Table~\ref{tab:maskrcnn1x} and Table~\ref{tab:maskrcnn3x} in different training settings. In Table~\ref{tab:maskrcnn1x}, we train Mask-RCNN for 12 epoch and resize the image to $800\times 1333$ while In Table~\ref{tab:maskrcnn3x}, we train the model for 36 epochs and resize the training images to different scales for data augmentation. We can see that the QuadTree attention obtains consistently better performance than other methods.

\begin{table}[]
\centering
\begin{tabular}{l|lllllll}
              & \#Params & AP   & AP$_{50}$ & AP$_{75}$ & AP$_S$ & AP$_M$ & AP$_L$ \\ \hline
QuadTree-B-b0 & \textbf{23.4}     & \textbf{42.4} & \textbf{64.5}      & \textbf{45.9}      & \textbf{38.9}   & \textbf{61.6}   & \textbf{41.6}  \\ \hline
QuadTree-B-b1 & \textbf{33.3}     & \textbf{46.4} & \textbf{68.6}      & \textbf{50.7}      & \textbf{41.9}   & \textbf{65.6}   & \textbf{44.7}   \\ \hline
Swin-T~\citep{liu2021swin}        & 47.8     & 46.0 & 68.1      & 50.3      & 41.6   & 65.1   & 44.9   \\
Focal-T~\citep{yang2021focal}       & 48.8     & 47.2 & 69.4      & 51.9      & 42.7   & 66.5   & 45.9   \\
QuadTree-B-b2 & \textbf{44.8}     & \textbf{49.3} & \textbf{70.7}      & \textbf{53.9}      & \textbf{43.9}   & \textbf{67.6}   & \textbf{47.4}   \\ \hline
Swin-S~\citep{liu2021swin}        & \textbf{69.1}     & 48.5 & 70.2      & 53.5      & 43.3   & 67.3   & 46.6   \\
Focal-S~\cite{yang2021focal}       & 71.2     & 48.8 & \textbf{70.5}      & 53.6      & 43.8   & 67.7   & 47.2   \\
QuadTree-B-b3 & 70.0     & \textbf{49.6} & 70.4      & \textbf{54.2}      & \textbf{44.0}   & \textbf{67.7}   & \textbf{47.5  }
\end{tabular}
\caption{Object detection results on COCO val2017 with Mask-RCNN training with 36 epochs and multi-scale data argumentation strategy. We use PVTv2 backbone and replace the reduction attention with quadtree attention.}\label{tab:maskrcnn3x}
\end{table}




\begin{figure}
\begin{subfigure}{.5\textwidth}
  \centering
  \includegraphics[width=.95\linewidth]{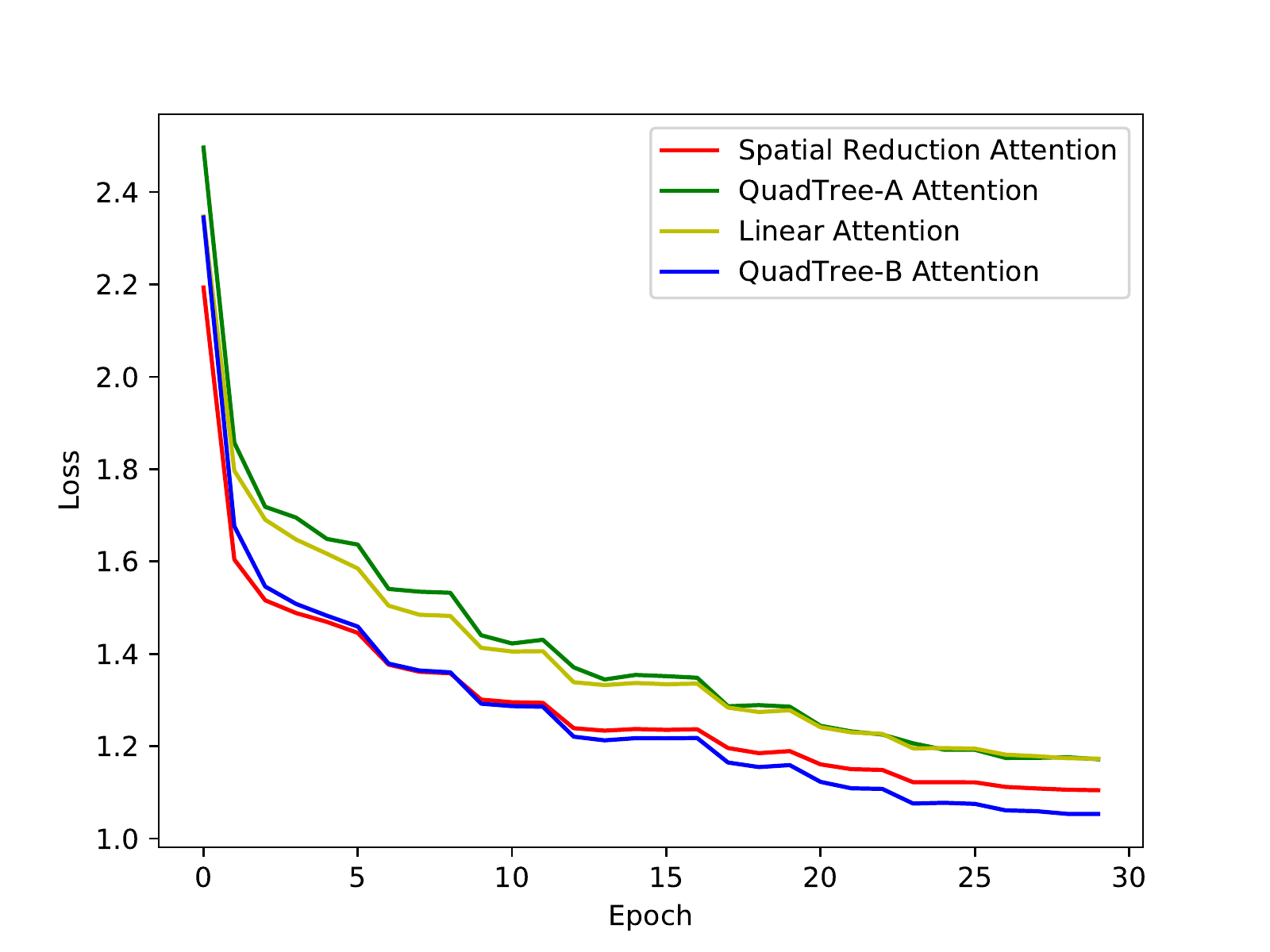}  
  \caption{Training Loss}
  \label{fig:sub-first}
\end{subfigure}
\begin{subfigure}{.5\textwidth}
  \centering
  \includegraphics[width=.95\linewidth]{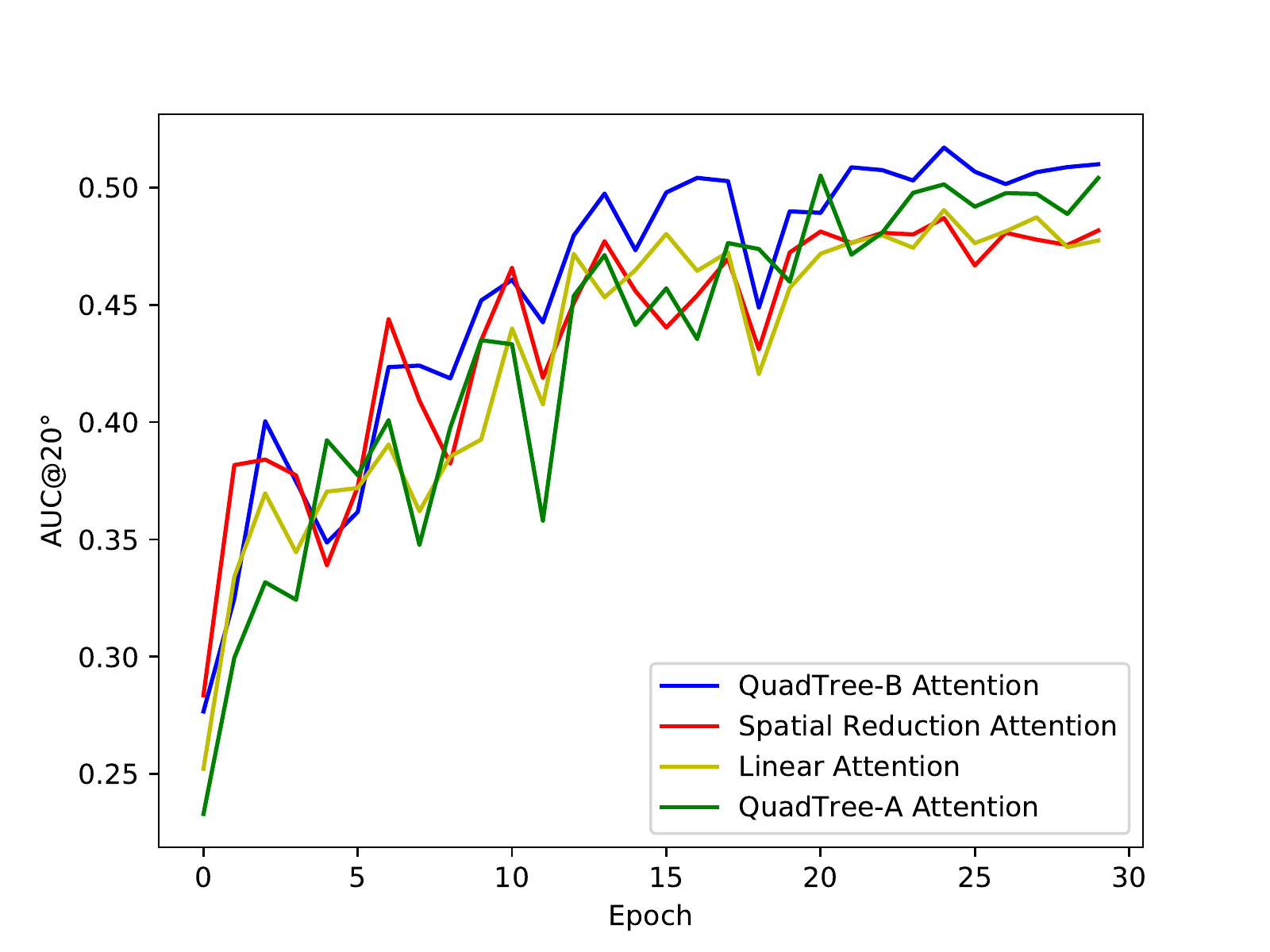}  
  \caption{AUC@20$^{\circ}$ }
  \label{fig:sub-second}
\end{subfigure}
\caption{Loss and AUC@20$^{\circ}$  of image matching.}
\label{fig:loss}
\end{figure}

\begin{figure}
\begin{subfigure}{.5\textwidth}
  \centering
  \includegraphics[width=.95\linewidth]{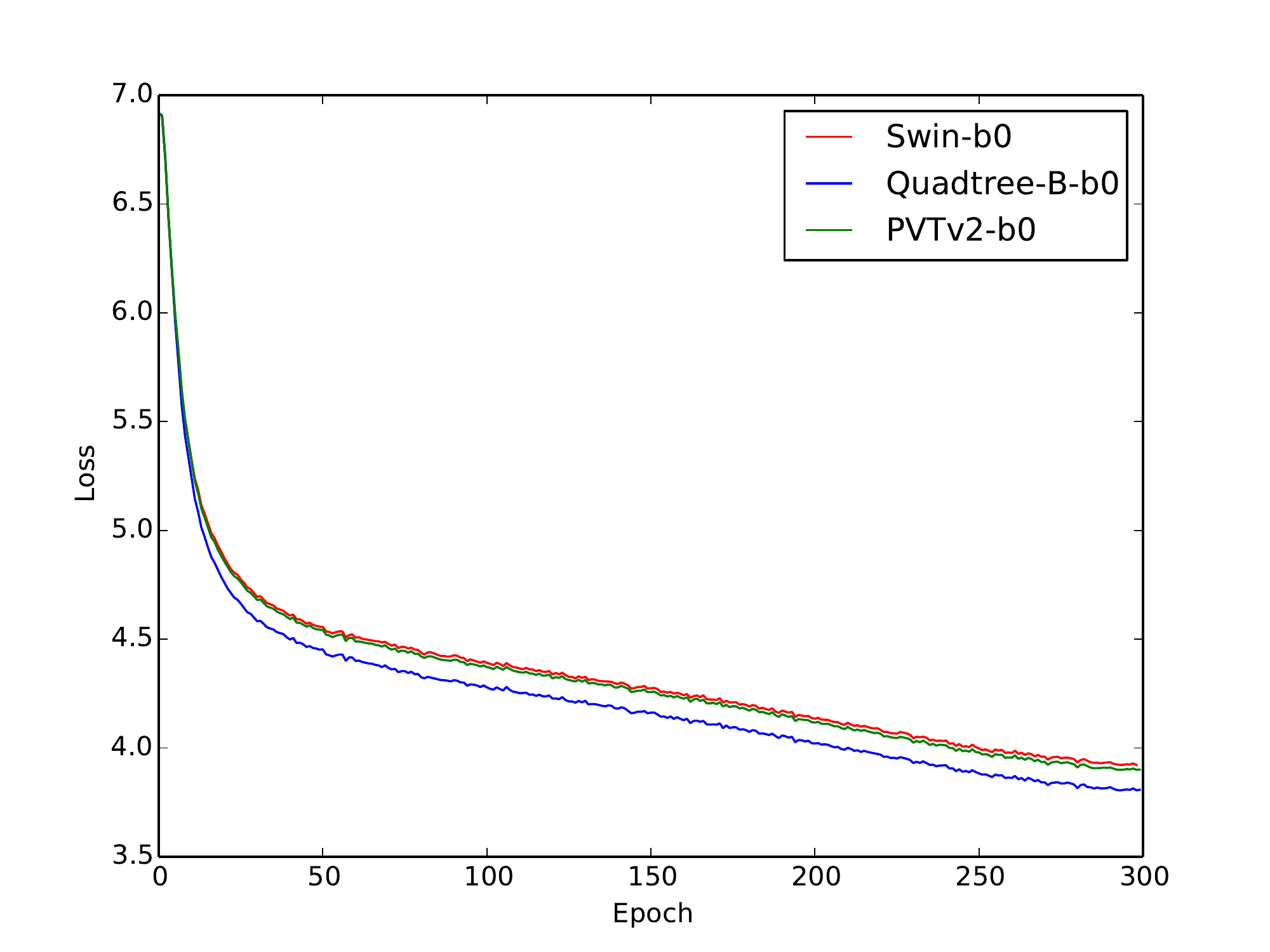}  
  \caption{Training Loss}
  \label{fig:sub-first}
\end{subfigure}
\begin{subfigure}{.5\textwidth}
  \centering
  \includegraphics[width=.95\linewidth]{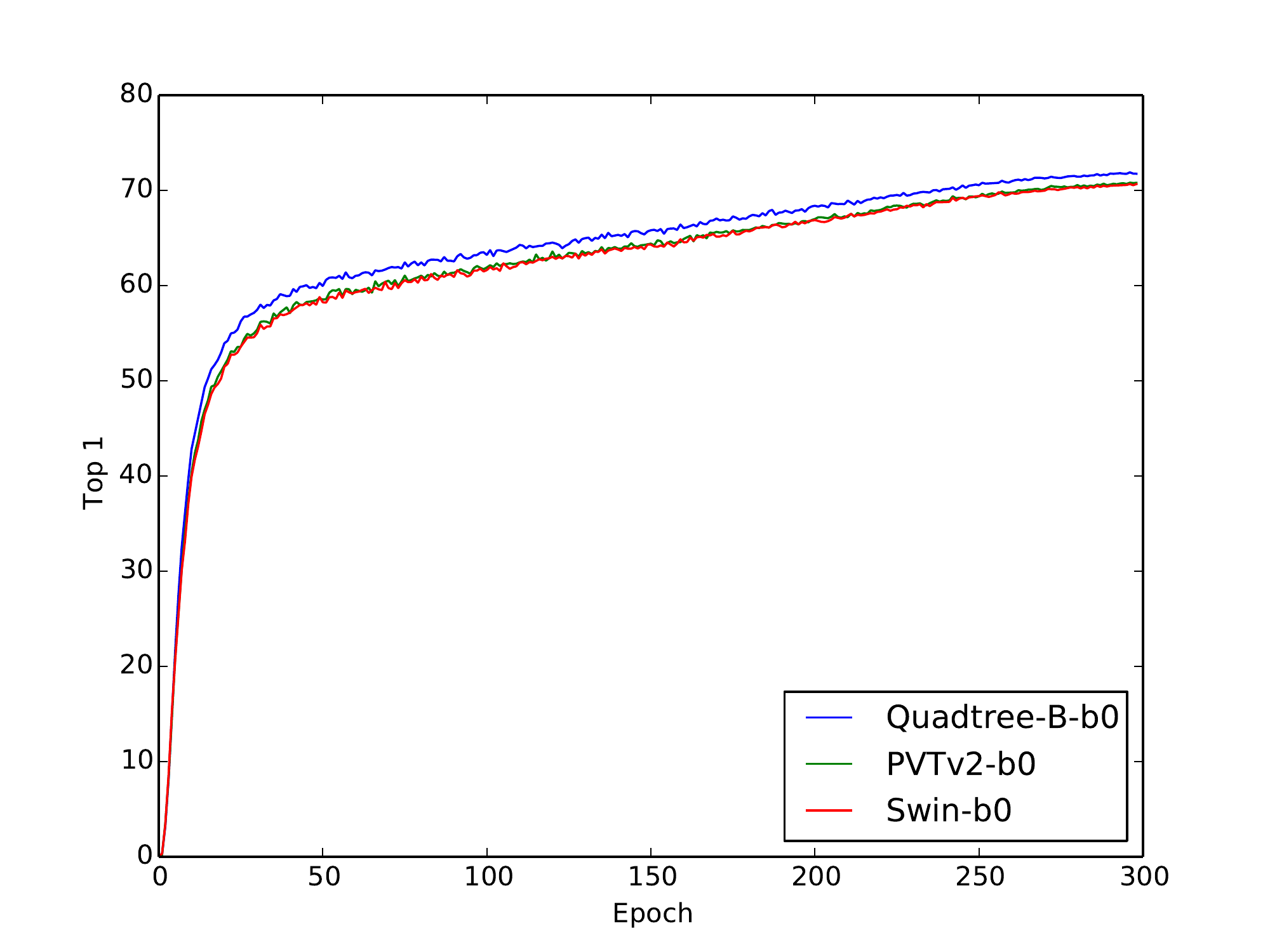}  
  \caption{Validation Accuracy}
  \label{fig:sub-second}
\end{subfigure}
\caption{Loss and top 1 accuracy of image classification for PVTv2-b0 archtecture.}
\label{fig:pvt_loss}
\end{figure}

\section{Training loss}
\textbf{Feature matching.} We plot the training loss and validation performance for LoFTR-lite in Figure~\ref{fig:loss} for different efficient transformers, including spatial reduction (SR) transformer~\citep{wang2021pyramid}, linear transformer~\citep{katharopoulos2020transformers}, our Quadtree-A, and Quadtree-B transformers. We can see quadtree-B transformer obtains consistently lower training loss and higher performance over other three transformers. In addition, it is also noted that the spatial reduction (SR) transformer has lower training but worse AUC@20$^{\circ}$  than QuadTree-A attention, which indicates that it cannot generalize well.

\textbf{Image classification}. We also show traning and validation curve for image classification task with respective to different attentions in Fig. \ref{fig:pvt_loss}. Compared with Swin Transformer~\citep{liu2021swin} and PVT~\citep{wang2021pyramid}, the loss of Quadtree attention is consistently lower and the top 1 accuracy is higher.

\section{Running time}
Currently, we only implement a naive CUDA kernel without many optimizations and it is not as efficient as the well-optimized dense GPU matrix operation. We test the running time of Retinanet under PVTv2-b0 architecture. For PVTv2-b0, The running time is 0.026s to forward one image and for Quadtree-b0, the running time is 0.046s for forwarding once. However, Quadtree-b0 has much lower memory usage than PVTv2-b0. Quadtree-b0 consumes about 339MB while PVTv2-b0 consumes about 574MB for one $800\times 1333$ image.

\section{Ablations}\label{apx:experiments}

\begin{figure}
\centering
\begin{subfigure}{.28\textwidth}
  \centering
  \includegraphics[width=.9\linewidth]{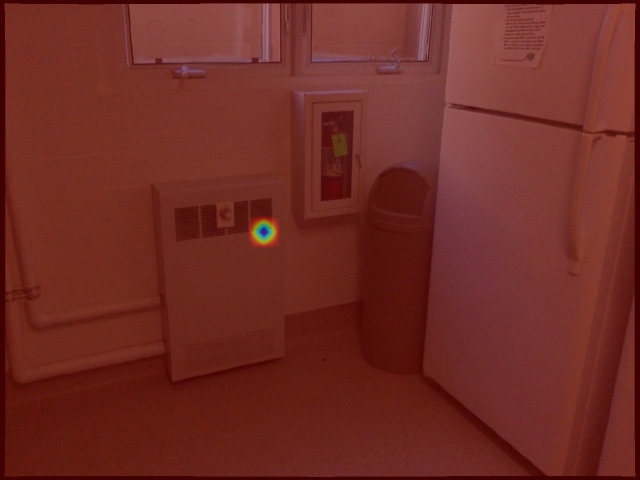}  
  \caption{Query image}
  \vspace*{2mm}
\end{subfigure}
\begin{subfigure}{.28\textwidth}
  \centering
  \includegraphics[width=.9\linewidth]{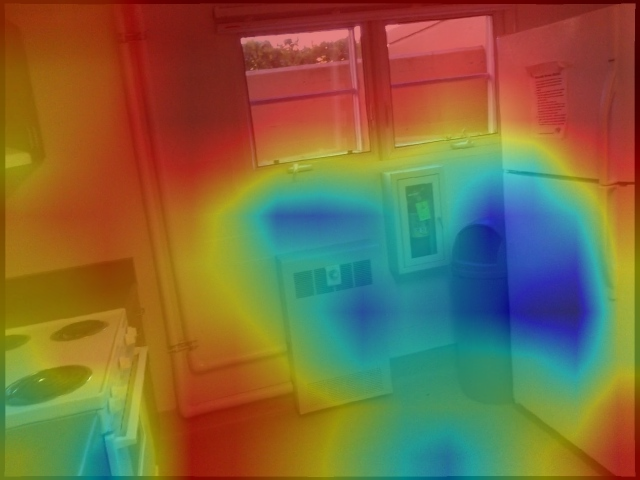}  
  \caption{SR attention}
  \vspace*{2mm}
\end{subfigure}
\begin{subfigure}{.28\textwidth}
  \centering
  \includegraphics[width=.9\linewidth]{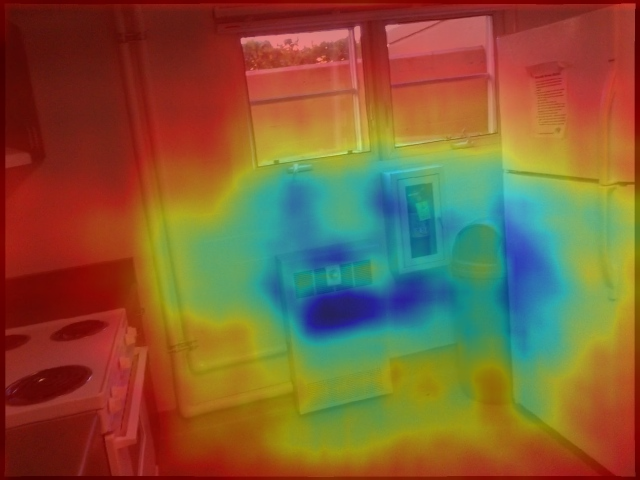}  
  \caption{Linear attention}
  \vspace*{2mm}
\end{subfigure}

\begin{subfigure}{.48\textwidth}
  \centering
  \includegraphics[width=.95\linewidth]{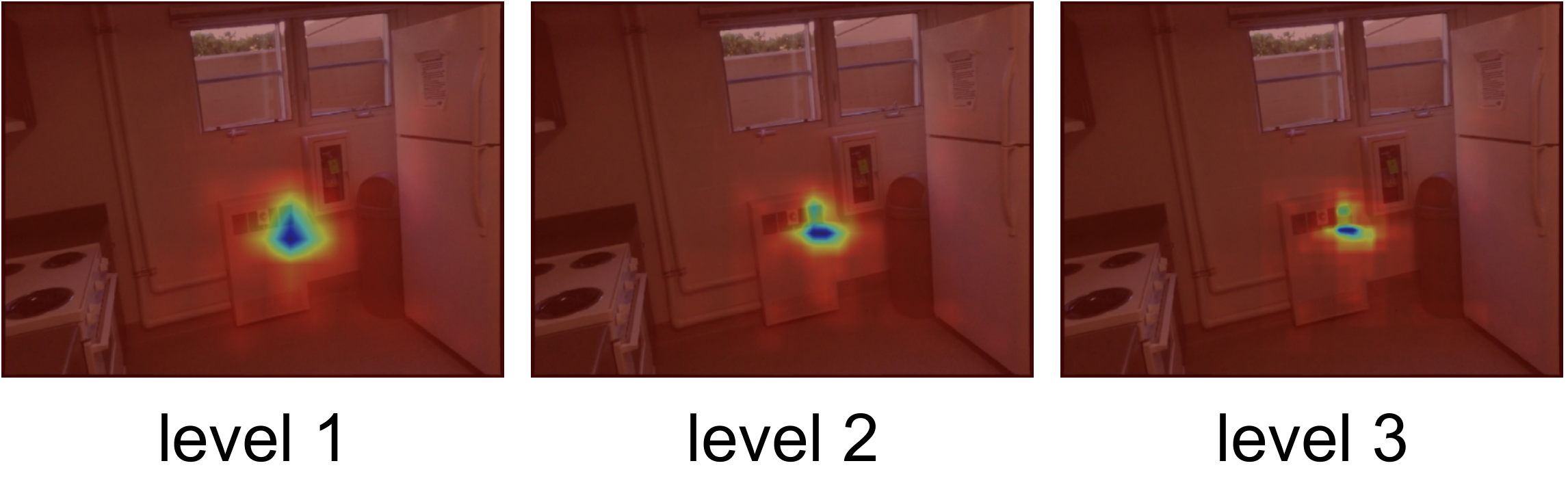}
  \caption{Quadree-A}
  \label{fig:sub-first}
\end{subfigure}
\begin{subfigure}{.48\textwidth}
  \centering
  \includegraphics[width=.95\linewidth]{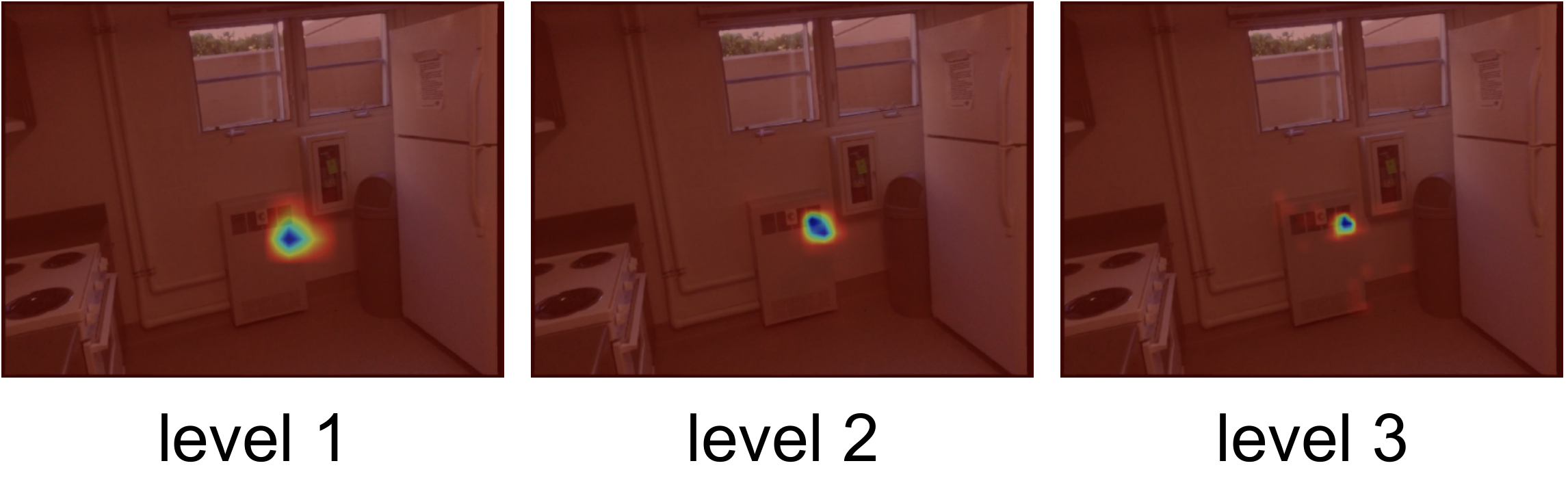}
  \caption{Quadtree-B}
  \label{fig:sub-second}
\end{subfigure}
\caption{Score map visualization of different attention methods for one patch in the query image. The first row shows score maps of spatial reduction attention and linear attention. The second row shows score maps of QuadTree-A and QuadTree-B at different levels, and the left image is the coarsest level, while the right image is the finest level for both sub-figures.}
\label{fig:score_vis}
\end{figure}

\begin{table}[]
\centering
\small
\begin{tabular}{llll|lll}
           & \multicolumn{3}{l|}{ImageNet}      & \multicolumn{3}{l}{COCO (RetinaNet)} \\ \hline
           & Param. (M) & Flops (G) & Top1 (\%) & AP         & AP$_{50}$       & AP$_{75}$       \\ \hline
Swin-T~\citep{liu2021swin}       & \textbf{29}        & \textbf{4.5}       & 81.3      & 42.0       &     \textbackslash       &       \textbackslash     \\
Focal-T~\citep{yang2021focal}       & \textbf{29}        & 4.9       & \textbf{82.2}      & 43.7       &     \textbackslash       &       \textbackslash     \\
Quadtree-B & 30        & 4.6       & \textbf{82.2 }     & \textbf{44.6}         & \textbf{65.8}         & \textbf{47.7}        
\end{tabular}
\caption{Comparison under Swin-T settings in image classification and object detection.}\label{tab:swin}
\end{table}

\textbf{QuadTree-A vs QuadTree-B}. QuadTree-B architecture consistently outperforms QuadTree-A in feature matching, image classification, and detection experiments. We analyze its reason as shown in Figure~\ref{fig:score_vis}, where (d) and  (e) show the attention score maps of QuadTree-A and QuadTree-B at different levels for the same point in the query image shown in (a). It is clear that the QuadTree-B has more accurate score maps, and is less affected by the inaccuracy in coarse level score estimation. We further visualize the attention scores of spatial reduction (SR) attention~\citep{wang2021pyramid} and linear transformer~\citep{katharopoulos2020transformers} in (b) and (c). We can see that SR attention and linear transformer attend the query token on large unrelated regions due to the loss of fine-grained information. In contrast, our quadtree transformer focus on the most relevant area. 

\begin{table}[]
\centering
\begin{tabular}{lll|lll}
               & \multicolumn{2}{l|}{ImageNet} & \multicolumn{3}{l}{COCO (RetinaNet)} \\ \hline
               & Flops (G)         & Top 1(\%)         & AP        & AP$_{50}$      & AP$_{75}$     \\ \hline
Quadtree-B-b2     & 4.3           &82.6          & 44.9      & 66.2        & 47.7       \\
Quadtree-B-b2+MPE & 4.3           & \textbf{82.7}          & \textbf{46.2}      & \textbf{67.2}        & \textbf{49.5}     
\end{tabular}
\caption{Ablation on multiscale position encoding.}
\end{table}

\textbf{Comparison with Swin Transformer and Focal Transformer.} We compare with Swin Transformer and Focal Transformer in Table. \ref{tab:swin} using the released codes. We replace the corresponding attention in Swin Transformer with Quadtree-B attention. Our method obtains 0.9\% higher top 1 accuracy than Swin Transformer and 2.6\% higher AP in object detection. Compared with Focal transformer, quadtree attention achieve the same top 1 accuracy in classification with fewer flops, and 0.9\% higher AP in object detection.

\textbf{Multiscale position encoding.} We compare our method with or without multiscale position encoding (MPE). For Quadtree-B-b2 model, MPE can bring an improvement of 1.3 on object detection.

\begin{table}
\centering
\small
\parbox{.48\linewidth}{
\centering
\begin{tabular}{l|lll}
     & AP   & $AP_{50}$ & $AP_{75}$ \\ \hline
$K=1$  &37.3 &57.2 &39.4       \\ \hline
$K=8$  &38.0 &58.2 &40.4       \\ \hline
$K=16$ &38.4 &58.7 &41.1     \\ \hline
$K=32$ &\textbf{38.5} &\textbf{58.8} &\textbf{41.1}    
\end{tabular}
\caption{The performance of QuadTree-B under different $K$ in object detection.}\label{tab:detection_k}
}
\hfill
\parbox{.48\linewidth}{
\centering
\small
\begin{tabular}{l|lll}
     & AUC@5$^{\circ}$     & AUC@10$^{\circ}$    & AUC@20$^{\circ}$    \\ \hline
$K=1$  & 15.7 & 32.3 & 48.9 \\ \hline
$K=4$  & 16.2 & 33.3 & 50.8 \\ \hline
$K=8$  & 17.4 & 34.4 & 51.6 \\ \hline
$K=16$ & \textbf{17.7} & \textbf{34.6} & \textbf{51.7}
\end{tabular}
\caption{The performance of QuadTree-B under different $K$ in feature matching}\label{tab:loftr_k}
}
\end{table}

\textbf{Top $K$ numbers.} Table~\ref{tab:detection_k} and Table~\ref{tab:loftr_k} shows the performance of QuadTree-B architecture with different value of $K$ for object detection and  feature matching respectively. The performance is improved when $K$ becomes larger and saturates quickly. This indicates only a few tokens with high attention scores should be subdivided in the next level for computing attentions.

\end{document}